\documentclass{article}

\usepackage{txie_icml}
\usepackage{natbib}
\usepackage{bbm}
\usepackage{microtype}
\usepackage{graphicx}
\usepackage{subcaption}
\usepackage{booktabs} % for professional tables
\usepackage{colortbl}
\usepackage{hyperref}
\usepackage{lipsum}

\hypersetup{
    colorlinks=true,
    }
\usepackage[accepted]{icml2024}

\usepackage{pifont}
\usepackage[capitalize,noabbrev]{cleveref}
\crefname{figure}{Fig.}{Figs.}
\crefname{table}{Tab.}{Tabs.}

\usepackage{amsmath}
\usepackage{amssymb}
\usepackage{mathtools}
\usepackage{amsthm}
\usepackage[inline]{enumitem}
\usepackage{adjustbox}
\usepackage{makecell}
\usepackage{enumitem}
\usepackage{tikz}
\newcommand{\blackcircle}[1]{%
\tikz[baseline=(char.base),baseline=-0.7ex]{\node[shape=circle,fill=black,text=white,inner sep=0.5pt,font=\scriptsize] (char) {#1};}%
}

\newcommand{\ourmethod}{\textbf{GST}\xspace}

\theoremstyle{plain}

\theoremstyle{definition}

\theoremstyle{remark}
\newtheorem{principle}{Principle}[section]

\usepackage[textsize=tiny]{todonotes}

\usepackage{xspace}
\usepackage[flushleft]{threeparttable}
\usepackage{tablefootnote}
\usepackage{multirow}

\usepackage[ruled,linesnumbered,algo2e]{algorithm2e}
\renewcommand{\algorithmicrequire}{\textbf{Input:}}  % Use Input in the format of Algorithm
\renewcommand{\algorithmicensure}{\textbf{Output:}} % Use Output in the format of Algorithm
\SetKwInOut{Input}{Input}\SetKwInOut{Output}{Output}
\SetKwComment{Comment}{$\triangleright$\ }{}
\usepackage{xcolor}

\definecolor{ForestGreen}{RGB}{34,139,34}
\definecolor{myyellow}{RGB}{181, 181, 27}
\newcommand{\greencheck}{{\color{ForestGreen}\cmark}}

\newcommand{\redcheck}{{\color{red}\xmark}}
\icmltitlerunning{Boosting Graph Sparse Training via Semantic and Topological Awareness}

\begin{document}

\twocolumn[
\icmltitle{Two Heads Are Better Than One:\\ Boosting Graph Sparse Training via Semantic and Topological Awareness}

% It is OKAY to include author information, even for blind
% submissions: the style file will automatically remove it for you
% unless you've provided the [accepted] option to the icml2024
% package.

% List of affiliations: The first argument should be a (short)
% identifier you will use later to specify author affiliations
% Academic affiliations should list Department, University, City, Region, Country
% Industry affiliations should list Company, City, Region, Country

% You can specify symbols, otherwise they are numbered in order.
% Ideally, you should not use this facility. Affiliations will be numbered
% in order of appearance and this is the preferred way.
\icmlsetsymbol{equal}{$\dag$}
% \icmlsetsymbol{equal}{$\ddag$}

\begin{icmlauthorlist}
\icmlauthor{Guibin Zhang}{tongji,equal}
\icmlauthor{Yanwei Yue}{tongji,equal}
\icmlauthor{Kun Wang}{ustc}
% Junfeng Fang, Yongduo Sui, Kai Wang, Yuxuan Liang, Dawei Cheng, Shirui Pan, Tianlong Chen
\icmlauthor{Junfeng Fang}{ustc}
\icmlauthor{Yongduo Sui}{ustc}
\icmlauthor{Kai Wang}{nus}
\icmlauthor{Yuxuan Liang}{ust}
%\icmlauthor{}{sch}
\icmlauthor{Dawei Cheng}{tongji}
\icmlauthor{Shirui Pan}{grif}
\icmlauthor{Tianlong Chen}{mit}
%\icmlauthor{}{sch}
%\icmlauthor{}{sch}
\end{icmlauthorlist}

\icmlaffiliation{tongji}{Tongji Univerity}
\icmlaffiliation{ustc}{University of Science and Technology of China}
\icmlaffiliation{nus}{National University of Singapore}
\icmlaffiliation{ust}{Hong Kong University of Science and Technology (Guangzhou Campus)}
\icmlaffiliation{grif}{Griffith University}
\icmlaffiliation{mit}{Massachusetts Institute of Technology}

\icmlcorrespondingauthor{Kun Wang}{wk520529@mail.ustc.edu.cu}
\icmlcorrespondingauthor{Tianlong Chen}{tianlong@mit.edu}

% You may provide any keywords that you
% find helpful for describing your paper; these are used to populate
% the "keywords" metadata in the PDF but will not be shown in the document
\icmlkeywords{Machine Learning, ICML}

\vskip 0.3in
]

% this must go after the closing bracket ] following \twocolumn[ ...

% This command actually creates the footnote in the first column
% listing the affiliations and the copyright notice.
% The command takes one argument, which is text to display at the start of the footnote.
% The \icmlEqualContribution command is standard text for equal contribution.
% Remove it (just {}) if you do not need this facility.

%\printAffiliationsAndNotice{}  % leave blank if no need to mention equal contribution
\printAffiliationsAndNotice{\icmlEqualContribution} % otherwise use the standard text.

\begin{abstract}
Graph Neural Networks (GNNs) excel in various graph learning tasks but face computational challenges when applied to large-scale graphs. A promising solution is to remove non-essential edges to reduce the computational overheads in GNN. Previous literature generally falls into two categories: topology-guided and semantic-guided. The former maintains certain graph topological properties yet often underperforms on GNNs due to low integration with neural network training. The latter performs well at lower sparsity on GNNs but faces performance collapse at higher sparsity levels. With this in mind, we take the \underline{first} step to propose a new research line and concept termed \textbf{Graph Sparse Training} \textbf{(GST)}, which dynamically manipulates sparsity at the data level. Specifically, GST initially constructs a topology \& semantic anchor at a low training cost, followed by performing dynamic sparse training to align the sparse graph with the anchor. We introduce the \textbf{Equilibria Sparsification Principle} to guide this process, effectively balancing the preservation of both topological and semantic information. Ultimately, GST produces a sparse graph with maximum topological integrity and no performance degradation.
Extensive experiments on 6 datasets and 5 backbones showcase that GST \textbf{(I)} identifies subgraphs at higher graph sparsity levels ($1.67\%\sim15.85\%$$\uparrow$) than state-of-the-art sparsifiers, \textbf{(II)} preserves more key spectral properties, \textbf{(III)} achieves $1.27-3.42\times$ speedup in GNN inference and \textbf{(IV)} successfully helps graph adversarial defense and graph lottery tickets. %\footnote{Source code is available at \url{https://anonymous.4open.science/r/GST-0F15}}
\end{abstract}

\section{Introduction}
\vspace{-0.5em}
Graph Neural Networks (GNNs) \cite{wu2020comprehensive,zhou2020graph} have emerged as leading solutions for graph-related learning tasks~\cite{velickovic2017graph,hamilton2017inductive,zhang2018link,zhang2019inductive,ying2018hierarchical,zhang2018end}. The essence of GNNs lies in their iterative neighborhood \emph{aggregation} and \emph{update} processes. The former aggregates neighboring node embeddings via sparse matrix-based operations (\textit{e.g.}, {\fontfamily{lmtt}\selectfont \textbf{SpMM}} and {\fontfamily{lmtt}\selectfont \textbf{SDDMM}}), while the latter updates central nodes using dense matrix-based operations (\textit{e.g.}, {\fontfamily{lmtt}\selectfont \textbf{MatMul}})~\cite{fey2019fast,wang2019deep}.
%In practice, the corresponding computational operations are {\fontfamily{lmtt}\selectfont \textbf{SpMM}} for \emph{aggregation} and {\fontfamily{lmtt}\selectfont \textbf{MatMul}} for \emph{update}~\cite{hu2020featgraph,qiu2021optimizing}. 
{\fontfamily{lmtt}\selectfont \textbf{SpMM}} often accounts for the largest portion (50\%$\sim$70\%) of GNN's computational load~\cite{qiu2021optimizing,liu2023dspar}, mainly determined by the size of the graph input. However, large-scale graph inputs are commonplace in real-world applications \cite{wang2022searching,jin2021graph,zhang2024graph}, presenting significant computational challenges that impede feature aggregation in GNN training and inference. These issues sadly drop a daunting obstacle on the way toward GNNs' on-device deployment, especially in resource-constrained environments.

\begin{figure}[t]
\centering
\includegraphics[width=1.0\columnwidth]{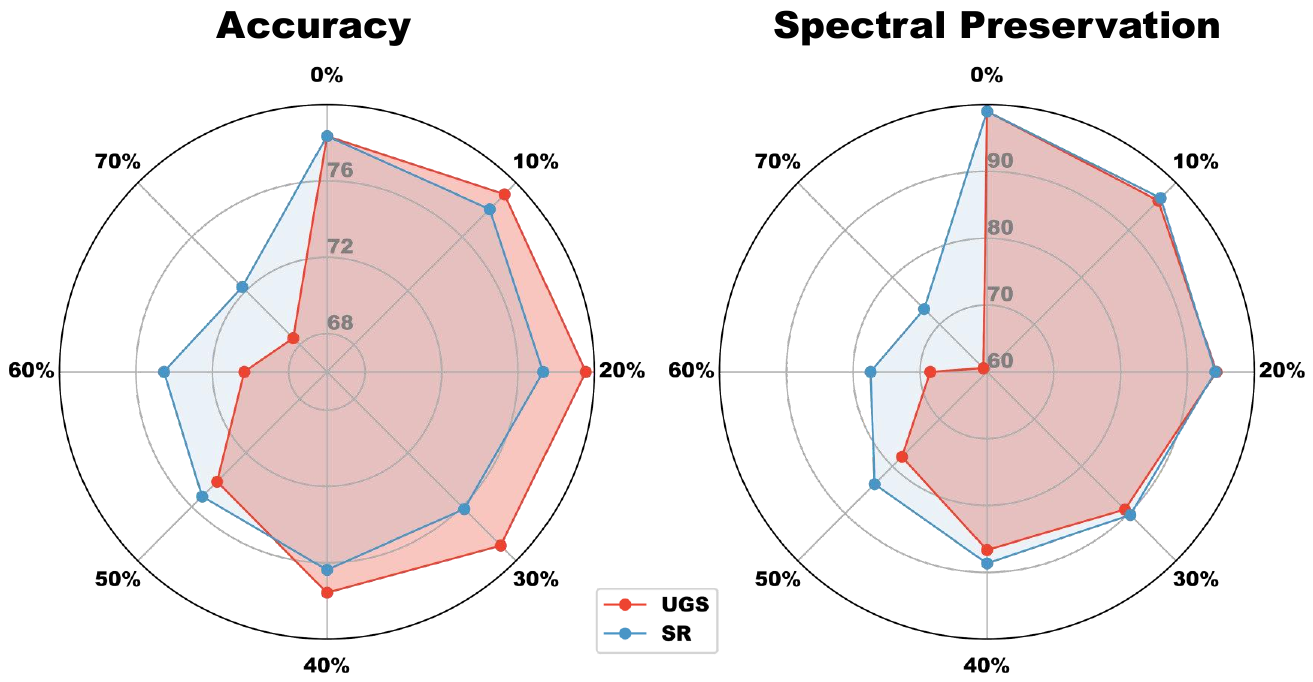}
\vspace{-0.6cm}
\caption{Graph sparsifier (UGS \& Spectral Radius) comparison on Ogbn-Proteins using 3-layer GraphSAGE at varying graph sparsity levels $\{10\%,20\%,\cdots,60\%\}$. (\textbf{\textit{Left}}) ROC-AUC score after different levels of sparsification. (\textbf{\textit{Right}}) The spectral preservation ratio\footnotemark of the obtained sparse graph.
}
\vspace{-2em}
\label{fig:intro}
\end{figure}

\footnotetext{Following~\cite{liu2023dspar}, we define the spectral preservation ratio as the relative error of the top-200 eigenvalues, represented as $\sum_{i=1}^{200}\frac{|\lambda_i - \lambda_i'|}{\lambda_i}$, where $\lambda_i$ and $\lambda_i'$ denote the $i$-th eigenvalue of the original and sparse graphs, respectively.}

Given that {\fontfamily{lmtt}\selectfont \textbf{SpMM}} dominates the computational load of GNNs, and the execution time of such sparse operation is proportional to the graph's edge count~\cite{liu2023dspar}, previous approaches to mitigate this inefficiency primarily concentrated on graph sparsification by discarding non-essential edges. They generally fall into two categories:

$\blacksquare$ The first research line involves calculating \textbf{topology-guided edge scores} for removing the less significant ones, such as resistance effective~\cite{spielman2008graph}, eigenvalue~\cite{batson2013spectral}, pairwise distances betweenness~\cite{david1995algorithms}, size of all cuts~\cite{benczur1996approximating} and node degree distributions~\cite{eden2018provable}. 
%Like some weight pruning at initialization (termed {\fontfamily{lmtt}\selectfont \textbf{PaI}}) methods, topology-guided pruning shares a similar concept. However, the key distinction lies in the latter's focused attention on the structural characteristics within the fixed input graph. 
$\blacksquare$ The other primarily relies on \textbf{semantic-based scores} derived from gradients, momentum, magnitude, \textit{etc.}, during GNN training, utilizing methods such as trainable masks~\cite{chen2021unified,wang2023adversarial,wang2023brave}, probabilistic sampling~\cite{zheng2020robust,luo2021learning} or meta-gradients~\cite{wan2021edge} to score and prune non-essential edges.

Scrutinizing the existing graph \textit{sparsifiers}, they focus on either topology or semantics,
thereby suffer from inherent limitations correspondingly:

\vspace{-0.7em}
\begin{itemize}[leftmargin=*,itemsep=-0.1em]
    \item[\ding{224}] \textbf{\textit{Limited Performance.}} Though theoretically capable of preserving specific spectral properties~\cite{batson2013spectral}, topology-guided sparsifiers overlook the rich graph features, GNN training dynamics, and downstream tasks, resulting in suboptimal performance when integrated with GNN~\cite{luo2021learning}. Semantic-guided sparsifiers dynamically assess edge importance during GNN training, while they often suffer from significant performance collapse when targeting high graph sparsity, due mostly to detrimental impact on the overall connectivity of the graph~\cite{hui2022rethinking, wang2023searching}. 

    % \vspace{-0.3em}
    \item[\ding{224}] \textbf{\textit{Empirical Observations.}} We apply a typical semantic-based sparsifier, UGS~\cite{chen2021unified} and a topology-based sparsifier, Spectral Radius (SR)~\cite{chan2016optimizing,karhadkar2022fosr} on Ogbn-Proteins~\cite{hu2020open}. From Fig.~\ref{fig:intro}, we observe that: \blackcircle{1} UGS maintains GNN performance well at lower sparsity levels (0\%$\sim$30\%), yet encounters a dramatic performance drop (over 5.5\%$\downarrow$) at higher sparsity (50\%$\sim$), accompanied by significant spectral preservation loss; \blackcircle{2} SR consistently fails to match GNN performance on sparse graphs with that on dense graphs,  though its performance deterioration is gradual, with a milder spectral preservation loss.
\end{itemize}
\vspace{-0.5em}

The aforementioned observations prompt questions about graph sparsifiers: \textit{Can we ideally leverage the strengths and mitigate the shortcomings of both research lines?} Going beyond this, considering topology-guided methods are inherently ahead of and independent of training, while semantic-guided ones are closely intertwined with GNN optimization, their integration is naturally incongruent. We further question that: \textit{How can we effectively combine topology- and semantic-aware sparsification, embodying the principle that two heads are better than one?}

To this end, we take the \underline{first step} to explore dynamic \underline{graph-level} sparse training in both a semantic- and topology-aware manner.
We innovatively propose \textbf{Graph Sparse Training} (\textbf{GST}), which iteratively updates and maintains a sub-counterpart of the original graph during training. 
To highlight, GST explores the dynamic sparse issue at data-level for the \underline{first time}, opening a potential pathway for integrating spectral theory and dynamic training algorithms. Additionally, we introduce the \textbf{Equilibria Sparsification Principle} to guide the exploration process, which offers a new paradigm for future sparsifier development.

% More specifically, GST starts by training the GNN with the original dense graph for a few epochs, capturing the best semantic and topological information as an \textit{anchor graph}. It then prunes the graph to the target sparsity via one-shot pruning, resulting in a sparse graph with suboptimal performance~\cite{ma2021sanity,frankle2020pruning,chen2021unified}. With this as a startpoint, the GNN continues to train while iteratively updating the graph, \textit{i.e.}, pruning and growing an equal number of edges, to discover an optimal sparse structure. During each update, both semantic and topological importance are considered. Edges are regrown to explicitly \textit{minimize the difference between the current sparse graph and the anchor graph}, dynamically exploring the graph structure to preserve maximal semantic and topological information. Briefly put, our contributions can be summarized as:
% Our experiments across 6 graph datasets and 5 backbones confirm that our GST outperforms all semantic- and topology-based graph sparsification methods in achieving higher GNN performance under extreme graph sparsity levels.
More specifically, considering the inherent properties of the graph itself, GST starts with full graph training to build the dependable anchor, providing a well-aligned semantic and topological benchmark for later dynamic sparse training. Then, GST prunes towards the larger sparsity, resulting in a sparse graph with suboptimal performance~\cite{ma2021sanity,frankle2020pruning,chen2021unified}. Finally, the GNN continues to train while iteratively updating the graph, \textit{i.e.}, pruning and growing an equal number of edges, to discover an optimal sparse structure. During each update, we adhere to the \textbf{Equilibria Sparsification Principle} that prioritizes both semantic and topological significance, \textit{explicitly minimizing the discrepancy between the current sparse graph and the anchor graph}. We methodically explore the graph structure through GST, aiming to retain the maximum amount of semantic and topological information. Briefly put, our contributions can be summarized as:

\vspace{-1em}
\begin{itemize}[leftmargin=*]
    \setlength\itemsep{0.1em}
    \item In this work, we systematically review two graph sparsification research lines: \textbf{topology-guided} and \textbf{semantic-guided}. For the first time, we develop a novel framework that combines their strengths and explicitly preserves graph topology integrity to boost GNN performance maintenance at extreme graph sparsity levels.
    \item We introduce \textbf{Graph Sparse Training} (\textbf{GST}), an innovative pruning framework that manipulates sparsity at the data level,  exploring sparse graph structure with both semantic- and topology-awareness. Additionally, GST boasts high versatility, effectively aiding various mainstream tasks, including graph adversarial defense and graph lottery tickets. 
    \item Our extensive experiments on 6 datasets and 5 backbones demonstrate that GST \textbf{(I)} identifies subgraphs at higher graph sparsity levels ($1.67\%\sim15.85\%\uparrow$) compared to SOTA sparsification methods in node classification tasks, without compromising performance, \textbf{(II)} effectively preserves $\sim15\%$ more spectral properties, \textbf{(III)} achieves tangible $1.27-3.42\times$ GNN inference speedup, and \textbf{(IV)} successfully combat edge perturbation ($0.35\%\sim7.23\%$$\uparrow$) and enhances graph lottery tickets ($0.22\%\sim1.58\%$$\uparrow$).
\end{itemize}

% \vspace{-1.5em}
\section{Related Work}\label{sec:related_work}
\vspace{-0.4em}
% \subsection{Graph Sparsification}
% \vspace{-0.5em}
\textbf{Topology-based sparsifiers} are early attempts at graph lightweighting. Essentially, they use a theoretically inspired \underline{pre-defined} metric to score edge importance and prune those with lower scores. %G-spar~\cite{murphy1996finley} and 
SCAN~\cite{xu2007scan} assesses global connectivity importance using Jaccard similarity. L-spar~\cite{satuluri2011local} and Local Similarity~\cite{hamann2016structure} filter edges locally based on Jaccard similarity. \cite{spielman2008graph} suggested graph sampling via effective resistance, and~\cite{liu2023dspar} accelerated its computation with unbiased approximation. 
However, these methods, typically effective in traditional graph tasks (\textit{e.g.,} graph clustering)~\cite{chen2023demystifying}, struggle to maintain performance when applied to GNNs due to their unawareness of GNN training~\cite{luo2021learning}.

\textbf{Semantic-based sparsifiers} are more closely integrated with GNNs. They aim to \underline{dynamically} identify important edges via GNN training semantics. NeuralSparse~\cite{zheng2020robust} introduces a learning-based sparsification network to select $k$ edges per node. 
%SGCN~\cite{li2020sgcn} approaches sparsification as an optimization objective and employs ADMM to solve it. 
Meta-gradient sparsifier~\cite{wan2021graph} leverages meta-gradients to periodically remove edges. Additionally, Graph Lottery Ticket (GLT)~\cite{chen2021unified,wang2023adversarial,wang2022searching} offers a new paradigm for graph sparsification, which gradually prunes the graph to target sparsity using iterative magnitude pruning (IMP). However, these methods often struggle at higher graph sparsity levels, due to their disruption of the graph topology~\cite{hui2022rethinking}.

\vspace{-0.4em}
\begin{table}[ht!]
\scriptsize
\setlength\tabcolsep{0.8pt}
\resizebox{\columnwidth}{!}{\begin{tabular}{l|ccc|c}
\hline
 \textbf{Sprase Training} & {\textbf{ Target}} & \textbf{Semantic?} & \textbf{ Topology?} & \textbf{PRC$^\S$}  \\ \hline \hline
 {SNIP~\cite{lee2018snip}} & Sparse NN & \greencheck & \redcheck & \greencheck  \\
% {SET~\cite{mocanu2018scalable}} & Sparse NN & \greencheck & \redcheck & \greencheck  \\
% {SNFS~\cite{dettmers2019sparse}} & Sparse NN & \greencheck & \redcheck & \greencheck \\
{RigL~\cite{evci2020rigging}} & Sparse NN & \greencheck & \redcheck & \greencheck  \\
% {ITOP~\cite{liu2021we}} & Sparse NN & \greencheck & \redcheck & \greencheck  \\
% {DST~\cite{liu2020dynamic}} & Sparse NN & \greencheck& \redcheck & \redcheck \\
% {MEST~\cite{yuan2021mest}} & Sparse NN & \greencheck& \redcheck & \greencheck  \\
{IMDB~\cite{hoang2023revisiting}} & Sparse NN & \redcheck & \greencheck & \greencheck   \\
{DSnT~\cite{zhang2023dynamic}} & Sparse LLM & \greencheck & \redcheck & \greencheck \\
\textbf{GST (Ours)} & Sparse Graph & \greencheck & \greencheck & \greencheck   \\
 \hline %\hline 
 \multicolumn{5}{l}{
  \tiny$\S$ PRC: Prune Rate Control
 }
 
\end{tabular}}
\vspace{-1.8em}
\caption{Comparison among different sparsifiers. }
\label{tab:sparsifiers_simple}
\vspace{-1em}
\end{table}

% \vspace{-0.3em}
% \subsection{Dynamic Sparse Training}
% \vspace{-0.3em}
\textbf{Dynamic Sparse Training (DST)} is gaining attention as a method to reduce the computational cost in network training. It entails starting with a randomly sparsified network and periodically updating its connections~\cite{evci2020rigging}. Recent advancements have expanded DST's scope both theoretically~\cite{liu2021we,huang2023dynamic} and algorithmically~\cite{liu2020dynamic,zhang2023dynamic}. \textbf{Our GST fundamentally differs from DST in at least two aspects}. as shown in \Cref{tab:sparsifiers_simple}: (1) \textit{Research target}. While current DST focuses on sparsifying network parameters, GST innovatively explores the feasibility of the ``prune and regrow" paradigm in graph data. (2) \textit{Update methodology}. DST essentially aligns with semantic-guided methods, adjusting connections based on semantic information (gradient, momentum, \textit{etc.}) during network training. In contrast, GST considers both semantic and topological information of the graph. 
We place more discussions and comparisons with related work in Appendix~\ref{app:related_work}.

\section{Methodology}\label{sec:method}
% \vspace{-0.4em}
\subsection{Notations and Formulations}
\vspace{-0.4em}
\noindent \textbf{Notations.} Consider a graph $\mathcal{G}=\{\mathcal{V},\mathcal{E}\}$, where $\mathcal{V}$ represents the nodes and $\mathcal{E}$ the edges. The feature matrix of $\mathcal{G}$ is $\mathbf{X} \in \mathbb{R}^{N \times D}$ , with $N = |\mathcal{V}|$ being the total node count and each node $v_i \in \mathcal{V}$ having an $D$-dimensional feature vector $\boldsymbol{x}_i = \mathbf{X}[i,\cdot]$. $\mathbf{A} \in \{0,1\}^{N \times N}$ is the adjacency matrix, where $\mathbf{A}[i,j] = 1$ signifies an edge $(v_i, v_j) \in \mathcal{E}$. Let $f(\cdot; \mathbf{\Theta})$ represent a GNN with $\mathbf{\Theta}$ as its parameters, then $\mathbf{Z} = f(\{\mathbf{A}, \mathbf{X}\};\mathbf{\Theta})$ is the model output.
Note that $\mathcal{G}$ is alternatively denoted as $\{ \mathbf{A},\mathbf{X}\}$ here, and we will interchangeably use its two notations. Consider semi-supervised node classification, its objective function $\mathcal{L}$ is:
\vspace{-0.3em}
\begin{equation}
\mathcal{L} (\mathcal{G},\mathbf{\Theta}) = -\frac{1}{|\mathcal{V}_{\text{label}}|}\sum_{v_i \in \mathcal{V}_{\text{label}}} y_i\operatorname{log}(z_i),
\end{equation}
where $\mathcal{L}$ is the cross-entropy loss calculated over the labelled node set $\mathcal{V}_{\text{label}}$, and $y_i$ and $z_i$ denotes the label and prediction of $v_i$, respectively.

\noindent \textbf{Problem Formulation.} The target
of graph sparsification is to identify the sparsest subgraph $\mathcal{G}^{\text{sub}} = \{\mathbf{A}\odot\mathbf{M}_g,\mathbf{X}\}$, where $\mathbf{M}_g\in \{0,1\}^{N\times N}$ is a binary mask indicating edge removal if $\mathbf{M}_g[i,j]=0$ for $(v_i,v_j)\in\mathcal{E}$ and $\odot$ denotes element-wise product. Additionally, the subgraph $\mathcal{G}^{\text{sub}}$ should satisfy the condition that training the GNN on $\mathcal{G}^{\text{sub}}$ achieves performance comparable to that on the original graph $\mathcal{G}$. The objective is as follows:
\vspace{-0.5em}
\begin{equation}
\begin{gathered}
\mathop{\arg \max}_{\mathbf{M}_g} \;s_g = 1 - \frac{\|\mathbf{M}_g\|_0}{\|\mathbf{A}\|_0} \\
\operatorname{s.t.} \left|\mathcal{L}\left(\{ \mathbf{A},\mathbf{X}\},\mathbf{\Theta}\right)-\mathcal{L} \left(\{ \mathbf{A}\odot\mathbf{M}_g,\mathbf{X}\},\mathbf{\Theta}\right)\right| < \epsilon,
\end{gathered}
\end{equation}
% \vspace{-0.5em}
where $s_g$ is the graph sparsity, $\|\cdot\|_0$ counts the number of non-zero elements, and $\epsilon$ is the threshold for permissible performance difference. We define the \textit{extreme graph sparsity} as the maximal $s_g$ without compromising accuracy.

\begin{figure*}[!ht]
\setlength{\abovecaptionskip}{6pt}
\centering
\includegraphics[width=0.95\textwidth]{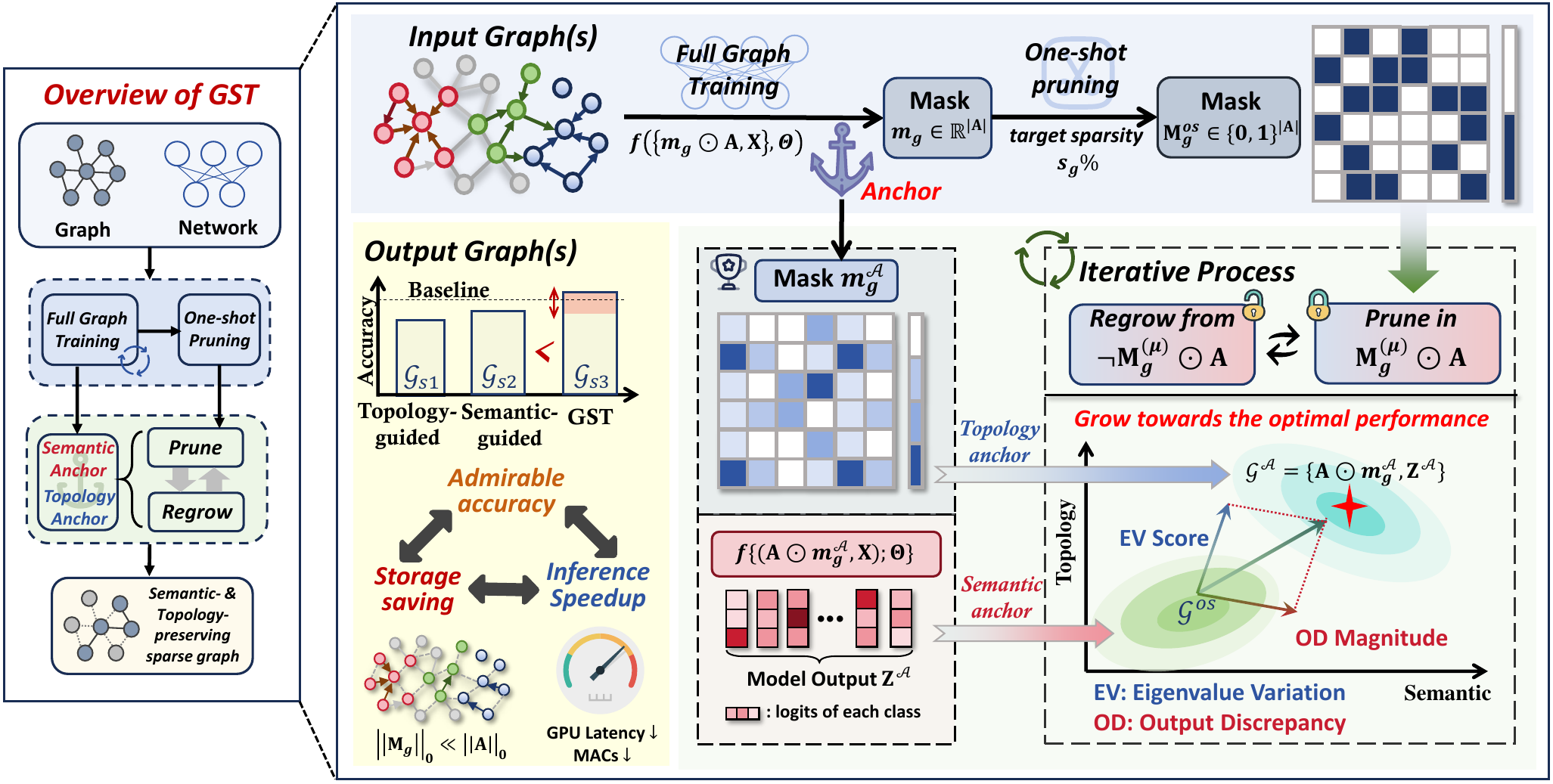}
\vspace{-1.0em}
\caption{(\textbf{\textit{Left}}) The overview of GST; (\textbf{\textit{Right}}) The detailed pipeline of GST. GST dynamically adjusts and updates the sparse graph, guided by an anchor graph from full-graph training, to optimize topological and semantic preservation, and finally yields a sparse subgraph at the desired sparsity along with admirable accuracy, storage saving, and inference speedup.} \label{fig:method}
\vspace{-0.9em}
\end{figure*}

\subsection{Framework Overview}
\vspace{-0.5em}
Fig.~\ref{fig:method} illustrates the overall workflow of GST. Starting with an initialized GNN and the adjacency matrix $\mathbf{A}$ as input, we first train the GNN together with a graph mask $\boldsymbol{m}_g \in \mathbb{R}^{|\mathbf{A}|}$ produced by a graph masker for limited epochs. During this phase, we capture the optimal topological and semantic information as the \textit{anchor graph} (Sec \ref{sec:anchor}). Subsequently, we apply one-shot pruning to $m_g$, reducing it to the desired sparsity $s_g\%$ and creating a sketched sparse graph. From this sparse base, we continue training, dynamically fine-tuning the sparse graph's structure to align semantically and topologically with the \textit{anchor graph} (Sec. \ref{sec:explore}). Through iterative refinement and exploration of graph structure, we ultimately achieve a sparse graph with maintained performance, reduced memory footprint, and faster inference speed (Sec. \ref{sec:criterion}).

\vspace{-0.5em}
\subsection{Pursuing Anchor Graph}\label{sec:anchor}
\vspace{-0.6em}
As outlined above, both topology-guided and semantic-guided sparsifiers typically derive guidance, explicitly or implicitly, from the original dense graph, striving to minimize their divergence from it~\cite{tsitsulin2023graph}. In line with the principle of efficiency in DST, several works such as Early-Bird (EB) and Graph EB have demonstrated that limited training can also construct high-quality benchmarks~\cite{achille2018critical,you2019drawing,you2022early}. Therefore, we propose conducting limited training on the original graph, \textit{i.e.}, full graph training, to capture an \emph{anchor} encompassing the original graph's topology and semantics. This approach provides natural ``instructions" for subsequent dynamic adjustment of the target sparse graph. 
To start with, we derive a (dense) graph mask $\boldsymbol{m}_g \in \mathbb{R}^{|\mathbf{A}|}$ with a parameterized graph masker $\Psi$:
% using the graph masker:
\vspace{-0.5em}
\begin{equation}
\boldsymbol{m}_g[i,j] = 
\begin{cases}
\Psi([x_i||x_j]),\; \text{if} (i,j)\in \mathcal{E},\\
0, \text{otherwise},
\end{cases}
\vspace{-0.4em}
\label{eq:masker}
\end{equation}
where $\boldsymbol{m}_g[i,j]$ denotes the edge score for $e_{ij}$, the graph masker $\Psi$ takes the concatenation of embeddings of $v_i$ and $v_j$ as input and output the edge score. Practically, we employ a 2-layer MLP for its implementation. We then co-optimize $\mathbf{\Theta}$, and $\Psi$ with the following loss function:
\vspace{-0.4em}
\begin{equation}\label{eq:loss}
\mathcal{L}_{\text{anchor}} = \mathcal{L} \left( \{\boldsymbol{m}_g \odot \mathbf{A}, \mathbf{X} \}, \mathbf{\Theta}\right),
\end{equation}
After the full graph training of $E$ epochs (practically $E \leq 200$), we select the optimal mask $\boldsymbol{m}_g^\mathcal{A}$ and model output $\mathbf{Z}^{\mathcal{A}}=f(\{\boldsymbol{m}^{\mathcal{A}}_g\odot\mathbf{A},\mathbf{X}\},\mathbf{\Theta}^\mathcal{A})$ from the epoch with the highest validation score, collectively termed as the \textit{anchor graph} $\mathcal{G}^\mathcal{A} = \{\boldsymbol{m}_g^\mathcal{A}\odot \mathbf{A},\mathbf{Z}^{\mathcal{A}}\}$. 
% Different from UGS~\cite{chen2021unified}, we do not impose $\ell_1$ regularization on $\boldsymbol{m}_g$ and $\boldsymbol{m}_\theta$ to reduce the usage of hyperparameters.
This process is rooted in an intuitive concept: \textit{although $\boldsymbol{m}_g$ and $\mathbf{\Theta}$ can be undertrained at this stage, the early training is capable of discovering vital connections and connectivity patterns}~\cite{achille2018critical,you2019drawing,you2022early}. Consequently, $\boldsymbol{m}_g^\mathcal{A}$ and $\mathbf{Z}^\mathcal{A}$ retain crucial properties for GNN training with the original graph $\mathcal{G}$. The anchor graph will be utilized in Sec.~\ref{sec:criterion}  to guide the exploration of sparse graph structure.

Given the target graph sparsity $s_g\%$, we zero the lowest-magnitude elements in $\boldsymbol{m}^{\mathcal{A}}_g$ w.r.t. $s_g$, and obtain its binarized version $\mathbf{M}_g^{\text{os}} \in \{0,1\}^{|{\boldsymbol{m}}_g^{\mathcal{A}}|}$. Such a sparse mask obtained via one-shot pruning is suboptimal~\cite{ma2021sanity,frankle2020pruning}, and we will start from this point and continually refine towards a more optimal graph structure.

\vspace{-0.5em}
\subsection{Dynamical Sparse Graph Training}\label{sec:explore}
\vspace{-0.3em}

With $\mathbf{M}^{\text{os}}_g$ at hand, we proceed to train the GNN model together with the fixed subgraph and the graph masker, denoted as $f(\{\boldsymbol{m}_g \odot \mathbf{M}_g^{\text{os}} \odot \mathbf{A}, \mathbf{X}\},\mathbf{\Theta})$, 
with the objective function similar
to Eq.~\ref{eq:loss}. We aim to gradually evolve the sparse graph structure toward both better topological and semantical preservation. To this end, we periodically reactivate the semantically and topologically significant edges in the pruned subgraph and substitute them for less important portions in the current subgraph within $D$ epochs.

We set the interval for each update (\textit{i.e.}, drop \ding{214} regrow) at $\Delta T$ epochs, with the total number of updates being $\lceil D / \Delta T \rceil$. We aim to develop a comprehensive criterion $\phi$ that evaluates the importance of an edge from both topological and semantic perspectives, which will guide the ``exchange of edges" between the current edges $\mathcal{E}_{(\mu)} = \mathbf{M}_g \odot \mathbf{A}$ and its complement $\mathcal{E}_{(\mu)}^{C} = \neg\mathbf{M}_g \odot \mathbf{A}$. Consider the update process between interval $\mu$ and $\mu+1$:
\begin{equation}\label{eq:choose}
\begin{aligned}
% \begin{cases}
\mathbf{M}^{(\text{prune})} = \operatorname{ArgTopK}\left(-\phi(\mathbf{M}_g^{(\mu)}\odot\mathbf{A}), \; \mathbf{\Upsilon}(\mu)\right), \\
\mathbf{M}^{(\text{regrow})} = \operatorname{ArgTopK}\left(\phi(\neg\mathbf{M}^{(\mu)}_g\odot\mathbf{A}), \; \mathbf{\Upsilon}(\mu)\right), 
% \end{cases}
\end{aligned}
\end{equation}
where $\operatorname{ArgTopK}(m,k)$ returns the indices of top-$k$ elements of matrix $m$, and $\mathbf{\Upsilon}(\cdot)$ is the update scheduler that controls the number of edges to be swapped at each update. The configuration of $\mathbf{\Upsilon}(\cdot)$ is detailed in Appendix \ref{app:parameter}. We then update the sparse graph as follows:
% \vspace{-0.5em}
\begin{equation}
\mathbf{M}_g^{(\mu+1)} = \left( \mathbf{M}_g^{(\mu)} \setminus \mathbf{M}_g^{(\text{prune})} \right) \cup \mathbf{M}_g^{(\text{regrow})}.
\label{eq:update}
\end{equation}
Then, in $(\mu+1)$-th interval, we continue training the GNN with the updated sparse graph $\mathbf{M}_g^{(\mu+1)} \odot \mathbf{A}$ for another $\Delta T$ epochs. This iterative process of updating and refining the sparse graph structure aims towards optimal performance. The remaining question now is: \textit{how do we design an ideal evaluation criterion $\phi$}?

\subsection{Topological \& Semantical Criterion Design}\label{sec:criterion}
\vspace{-0.4em}

To answer the question above, we introduce the Equilibria Principle to guide the design of graph pruning criteria, aiming to establish a new paradigm for sparsifier development:

\begin{principle}[\textit{\textbf{Equilibria Sparsification Principle}}]\emph{
Given a graph $\mathcal{G}=\{\mathbf{A},\mathbf{X}\}$ and target sparsity $s_g\%$, an ideal sparsifier $\mathcal{M}_g\in\{0,1\}^{|\mathbf{A}|}(\frac{\|\mathcal{M}_g\|_0}{\|\mathcal{A}\|_0}=s_g\%)$ and its resulting subgraph ${\mathcal{G}}^{\text{sub}}=\{\mathcal{M}_g\odot\mathbf{A},\mathbf{X}\}$ should satisfy the following condition:
\vspace{-0.5em}
\begin{equation}
\begin{aligned}
\mathop{\arg \min}_{\mathcal{M}_g} \mathbb{E}\;\left(\sum_{\mathcal{T}'\in\{\mathcal{T}\}}\mathcal{T}'(\mathcal{G}, \mathcal{G}^{\text{sub}}) + \sum_{\mathcal{S}'\in\{\mathcal{S}\}}\mathcal{S}'(\mathcal{G}, \mathcal{G}^{\text{sub}})\right),
% \textrm{s.t.}\quad(\frac{\|\mathcal{M}_g\|_0}{\|\mathcal{A}\|_0}=s_g\%)
\end{aligned}
\end{equation}
where $\mathbb{E}$ denotes the expectation, $\{\mathcal{T}\}$ denotes all possible metrics measuring the topological information difference between the sparse and original graphs (graph distance, spectral gap, \textit{etc.}), and $\{\mathcal{S}\}$ represents all possible metrics measuring semantic information differences (gradients, momentum, \textit{etc.}).}
\end{principle}

However, it is impractical to traverse and satisfy all possible metrics, so we provide two exemplified approaches for topology/semantics preservation metrics $\mathcal{T}$ and $\mathcal{S}$, and utilize them to guide the update process in Sec.~\ref{sec:explore}. 
% Recalling from Sec.~\ref{sec:anchor}, we noted the optimal graph mask $\boldsymbol{m}_g^\mathcal{A}$ and model output $\mathbf{Z}^\mathcal{A}$ as anchors for topology and semantics, respectively. We'll utilize these here.

% Here, we employ these to design the drop \& regrow criterion, both explicitly aimed at aligning the sparse graph with the original dense graph.
%$\boldsymbol{m}_g^{\mathcal{A}} \odot \mathbf{A}$.

\textbf{Topology Criterion.} Despite the myriad of edge scoring methods~\cite{tsitsulin2023graph}, we opt for \textbf{eigenvalue variation}, \emph{i.e.}, the relative error of all eigenvalues, as the metric for edge dropping/regrowing. This is because most topology-guided scores, including effective resistance~\cite{spielman2008graph}, spectral radius~\cite{costa2007characterization}, and graph curvature~\cite{tsitsulin2023graph,forman2003bochner}, rely partially or wholly on graph Laplacian eigenvalues. Ideally, we aim to identify a set of edges $\mathcal{E}'$ from $\mathcal{E}_{(\mu)}$ that minimally impact the graph Laplacian and a set $\mathcal{E}''$ from  $\mathcal{E}_{(\mu)}^C$  that maximally affect it $(|\mathcal{E}'| = |\mathcal{E}''| = \mathbf{\Upsilon}(\mu))$, thereby guiding the sparse graph structure towards restoring the topological properties of the anchor graph. The objective is as follows:
\vspace{-0.6em}
\begin{equation}\small
\begin{aligned}
\mathop{\arg \min}_{\mathcal{E}',\mathcal{E}''}\; \mathbb{E}\left(\mathcal{T}(\mathcal{G}^{\mathcal{A}}, \mathcal{G}^{(\mu+1)})\right), \mathcal{T}(\mathcal{G}_1,\mathcal{G}_2) = \sum_{i=1}^{N}\frac{|\lambda^{(1)}_i - \lambda^{(2)}_i|}{\lambda^{(1)}_i},
\end{aligned}
\end{equation}
where $\mathcal{G}^{(\mu+1)}=\{\mathcal{V},\mathcal{E}_{(\mu+1)}\}$ is the updated sparse graph after the $\mu$-th update, and we choose eigenvalue variation as the implementation for $\mathcal{T}$.
However, exhaustively evaluating all combinations of $(\mathcal{E}',\mathcal{E}'')$ is impractical. Therefore, we shift to assessing the impact of individual edges on the graph Laplacian as a measure of their importance:
\vspace{-0.4em}
\begin{equation}
\begin{aligned}
\phi^{\text{(topo)}}(e_{ij}) = \sum_{k=1}^{N} \frac{|\lambda_k(\mathcal{G}^{\mathcal{A}}) - \lambda_k(\mathcal{G}^{\mathcal{A}} \setminus e_{ij})|}{\lambda_k(\mathcal{G}^{\mathcal{A}})},
\end{aligned}
\end{equation}
where $\lambda_k(\mathcal{G}^{\mathcal{A}})$ denotes the $k$-th eigenvalue of the anchor graph, and $\lambda_k(\mathcal{G}^{\mathcal{A}} \setminus e_{ij})$ represents that after removing edge $e_{ij}$. For computational feasibility, we provide an approximate version, with detailed derivation in Appendix~\ref{app:eigen}:
\begin{equation}
\phi^{\text{(topo)}}(e_{ij}) = \left(\sum_{k=1}^{K} + \sum_{k=N-K}^{N}\right)\frac{\boldsymbol{m}^{\mathcal{A}}_{g,ij}a^{(k)}_ib^{(k)}_j}{\lambda_k(\mathcal{G}^{\mathcal{A}})a^{(k)T}b^{(k)}},
\label{eq:topo}
\end{equation}
where $a^{(k)}$ ($b^{(k)}$) is the left (right) eigenvector of the anchor graph's $k$-th eigenvalue. Notably, we only consider the top-$K$ and bottom-$K$ eigenvalues (practically, $K$=20) as they sufficiently reflect the overall properties of the graph.

\textbf{Semantic Criterion.} 
From the perspective of semantic preservation, we formulate the objective as follows:
\vspace{-0.4em}
\begin{equation}
\begin{gathered}
\mathop{\arg \min}_{\mathcal{E}',\mathcal{E}''}\; \mathbb{E}\left(\mathcal{S}(\mathcal{G}^{\mathcal{A}}, \mathcal{G}^{(\mu+1)})\right),\\ \mathcal{S}(\mathcal{G}^\mathcal{A},\mathcal{G}^{(\mu)}) = \mathbb{KL}\left(f(\mathcal{G}^{\mathcal{A}},\mathbf{\Theta}^{\mathcal{A}}),f(\mathcal{G}^{(\mu)},\mathbf{\Theta})\right),
\end{gathered}
\end{equation}
where $\mathbf{Z}^\mathcal{A}=f(\mathcal{G}^{\mathcal{A}},\mathbf{\Theta}^{\mathcal{A}})$ is the semantic anchor,  $f(\mathcal{G}^{(\mu)},\mathbf{\Theta})$ is the current model output, and $\mathcal{S}(\cdot,\cdot)$ employ the KL divergence~\cite{sun2022graph} to evaluate the model output discrepancies. Based on this, we propose the semantic criterion:
\vspace{-0.3em}
\begin{equation}
\phi^{(\text{sema})}(e_{ij}) = \left|\nabla_{e_{ij}}\mathcal{S}(\mathcal{G}^\mathcal{A},\mathcal{G}^{(\mu)})\right|,
\label{eq:sema}
\end{equation}
where edges with greater gradient magnitude after a single backpropagation step on output discrepancies are considered more valuable, otherwise the opposite.
\begin{figure*}[!ht]
\centering
\includegraphics[width=1\textwidth]{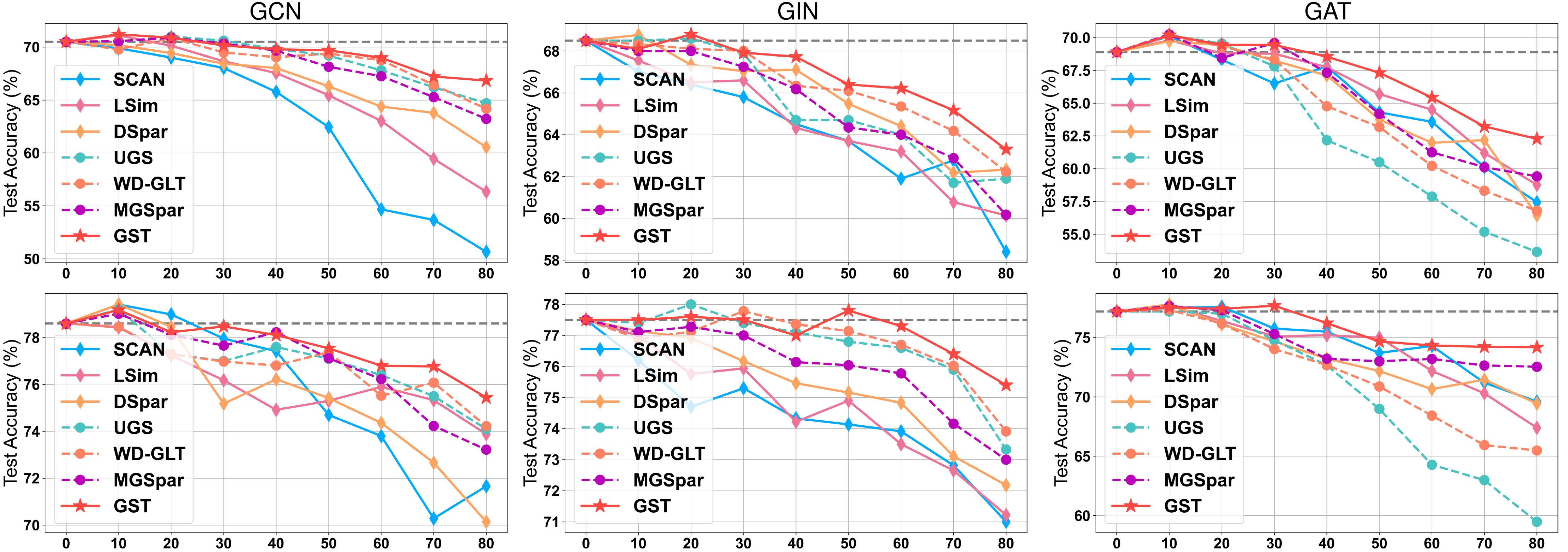}
\vspace{-2em}
\caption{Performance comparison of graph sparsification methods on Citeseer (\textbf{\textit{First Row}}) and PubMed (\textbf{\textit{Second Row}}) under different sparsity levels. The gray dashed line represents the original baseline performance.}
\vspace{-1em}
\label{fig:rq1_1}
\end{figure*}

\textbf{{Equilibria Combination.}} After selecting appropriate topological/semantic criteria, we combine these to form the final drop/regrow criterion, aligning with the equilibria principle:
\begin{equation}
\phi(e_{ij}) = \beta^s\cdot\phi^{\text{(sema)}}(e_{ij}) + \beta^t\cdot\phi^{\text{(topo)}}(e_{ij}),
\label{eq:combine}
\end{equation}
where $\beta^s$ and $\beta^t$ are corresponding scaling coefficients. At the end of interval $\mu$, we exchange edges with the highest $\phi$ in $\mathcal{E}_{(\mu)}^C$ and those with the lowest $\phi$ in $\mathcal{E}_{(\mu)}^C$, resulting in the updated sparse graph $\mathcal{G}^{(\mu+1)}$, as described in Sec.~\ref{sec:explore}.

\noindent \textbf{Model Summary.} Through such iterative exploration of graph structure, GST eventually achieves a sparse subgraph at the desired sparsity with no performance degradation. The overall algorithm framework is showcased in Algo.~\ref{alg:algo}, and its complexity analysis is presented in Appendix~\ref{app:complexity}. Importantly, GST, as a versatile concept, exhibits compatibility with various mainstream research lines, thereby providing substantial support for advancements in these areas, such as graph adversarial defense and graph lottery ticket, \textit{etc}. (discussed in Sec.~\ref{sec:rq4}).

\vspace{-0.6em}
\section{Experiments}
\vspace{-0.6em}
In this section, we conduct extensive experiments to answer the following research questions ($\mathcal{RQ}$): 
\vspace{-0.8em}
\begin{enumerate}[start=1,label={\bfseries $\mathcal{RQ}$\arabic*:},leftmargin=3em,itemsep=-1mm]
\item Can GST effectively find resilient sparse subgraphs?
\item How effective is GST in terms of preserving spectral information, \textit{i.e.}, eigenvalues?
\item Does GST genuinely accelerate the GNN inference?
\item Can GST serve as a universal operator?
\item How sensitive is GST to its key hyperparameters and components?
\end{enumerate}

\vspace{-0.9em}
\subsection{Experiment Setup}
\vspace{-0.4em}
\noindent \textbf{Dataset.} To exhaustively assess GST across various benchmarks and tasks, we select three popular graph datasets, including Cora, Citeseer, and PubMed~\cite{kipf2016semi}. For larger-scale graphs, we utilize Ogbn-ArXiv, Ogbn-Proteins, and Ogbn-Products~\cite{hu2020open}. 

\noindent \textbf{Backbones.} We evaluate GST under both transductive and inductive settings. For \textbf{transductive settings}, we select GCN~\cite{kipf2017semi}, GIN~\cite{xu2018powerful} and GAT~\cite{velivckovic2017graph} for full-batch training. For \textbf{inductive settings}, we select GraphSAGE~\cite{hamilton2017inductive} and ClusterGCN~\cite{chiang2019cluster}.

\noindent \textbf{Baselines.} We compare GST against two categories of graph sparsification methods:  (1) \textbf{topology-based sparsification}, including SCAN~\cite{xu2007scan}, Local Similarity (LSim)~\cite{satuluri2011local}, and DSpar~\cite{liu2023dspar}; (2) \textbf{semantic-based sparsification}, including UGS~\cite{chen2021unified}, meta-gradient sparsifier (MGSpar)~\cite{wan2021edge}, and WD-GLT~\cite{hui2022rethinking}.
More details on experiment setup can be found in Appendix~\ref{sec:app_setup}.

\vspace{-0.6em}
\subsection{GST Excels In Combating Sparsity ($\mathcal{RQ}$1)}
\label{sec:rq1}
\vspace{-0.5em}
We present the performance of GCN/GIN/GAT on Cora/Citeseer/PubMed in \Cref{fig:rq1_1,fig:rq2_2} and that of GraphSAGE/ClusterGCN on Ogbn-ArXiv/Proteins/Products in \Cref{tab:rq1_arxiv,tab:rq1_proteins,tab:rq1_products}. Each point in the figures represents the test accuracy/ROCAUC of the GNNs with the corresponding sparsifier at various levels of graph sparsity. Our observations are as follows:

\textbf{Obs.\blackcircle{1} GST is flexible and consistently outperforms other sparsifiers.} \Cref{fig:rq1_1,fig:rq2_2} demonstrate that GST \textbf{(I)} maintains GNN performance best at lower sparsity levels, such as on GIN+PubMed with $50\%$ graph sparsity, where other sparsifiers experienced a performance drop of $0.36\%\sim3.17\%$ compared to the baseline, whereas GST even showed $0.3\%$ improvement; \textbf{(II)} most effectively counters the adverse effects of sparsification at extreme sparsity levels, outperforming other sparsifiers by $1.23\%\sim5.39\%$ at $80\%$ graph sparsity.

% \vspace{-0.5cm}
\vspace{-0.2em}
\begin{table}[ht]\footnotesize
\centering
\caption{Performance comparison of sparsifiers at different sparsity levels ($10\%\rightarrow60\%$) on GraphSAGE/ClusterGCN with Ogbn-ArXiv. We report the
mean accuracy ± stdev of 3 runs. We shade the best-performing value in each column.}
% \vspace{0.4em}
\setlength{\tabcolsep}{1.80pt}
\resizebox{\columnwidth}{!}{
\begin{tabular}{l|cccccc}
\toprule
 % & \multicolumn{6}{c}{\textbf{GraphSAGE}}  \\ 
% \cmidrule(r){2-5}\cmidrule(r){6-9}
% \cmidrule(r){1-1}\cmidrule(r){2-7}%\cmidrule(r){5-7}
% \cmidrule(r){1-1}\cmidrule(r){2-5}
 % Ticket& \multicolumn{4}{c}{\textbf{Weight Sparsity}}  \\ 
\textbf{GraphSAGE}& $10\%$ & $20\%$ & $30\%$  & $40\%$ & $50\%$ & $60\%$ \\
% \midrule
\cmidrule(r){1-1}\cmidrule(r){2-7}%\cmidrule(r){5-7}

LSim & $69.22_{\pm0.11}$ &  $68.40_{\pm0.18}$ &  $66.15_{\pm0.22}$  &  $64.66_{\pm0.31}$  &  $61.07_{\pm0.23}$ &  $58.21_{\pm0.09}$ \\
DSpar &  $\cellcolor{gray!25}71.23_{\pm0.24}$&  \underline{$71.03_{\pm0.28}$} &  \underline{$68.50_{\pm0.33}$} &  $64.57_{\pm0.26}$  &  $62.79_{\pm0.66}$  &  $60.49_{\pm0.58}$\\

UGS &  $68.77_{\pm0.21}$&  $67.92_{\pm0.53}$ &  $66.30_{\pm0.27}$&  \underline{$66.57_{\pm0.18}$}  &  \underline{$65.72_{\pm0.14}$}  &  \underline{$63.40_{\pm0.36}$}\\

\ourmethod &  \underline{$71.12_{\pm0.23}$} &  \cellcolor{gray!25}$71.14_{\pm0.18}$  &  \cellcolor{gray!25}$71.46_{\pm0.37}$ &  \cellcolor{gray!25}$70.57_{\pm0.34}$ &  \cellcolor{gray!25}$68.02_{\pm0.58}$ &  \cellcolor{gray!25}$66.55_{\pm0.53}$\\
\midrule

\textbf{ClusterGCN}& $10\%$ & $20\%$ & $30\%$   & $40\%$ & $50\%$ & $60\%$ \\
% \midrule
\cmidrule(r){1-1}\cmidrule(r){2-7}%\cmidrule(r){5-7}
LSim & $67.27_{\pm0.54}$ &  $67.80_{\pm0.46}$  & $65.49_{\pm0.45}$ & $64.97_{\pm0.52}$ & $63.56_{\pm0.39}$
& $62.18_{\pm0.39}$\\
DSpar &  \underline{$68.75_{\pm0.75}$ }&  \underline{$68.40_{\pm0.69}$}  &  \underline{$66.72_{\pm0.56}$}  &  \underline{$66.09_{\pm0.71}$}
&  \underline{$65.45_{\pm0.51}$}
&  \underline{$63.72_{\pm0.50}$}\\
\ourmethod & \cellcolor{gray!25} $69.44_{\pm0.14}$ &  \cellcolor{gray!25}$69.21_{\pm0.24}$  &  \cellcolor{gray!25}$68.17_{\pm0.63}$ &  \cellcolor{gray!25}$68.02_{\pm0.49}$ &  \cellcolor{gray!25}$67.33_{\pm0.37}$
&  \cellcolor{gray!25}$65.98_{\pm0.50}$\\
\bottomrule
\end{tabular}}
\label{tab:rq1_arxiv}
\vspace{-1.5em}
\end{table}

\textbf{Obs.\blackcircle{2} GST resiliently scales to large-scale graphs.} As demonstrated in \Cref{tab:rq1_arxiv,tab:rq1_proteins,tab:rq1_products}, GST maintains robust performance when sparsifying large graphs under inductive setting. Specifically, on Ogbn-ArXiv with $40\%$ sparsity, GST+GraphSAGE/ClusterGCN experienced negligible performance losses of only $0.43\%$ and $0.48\%$, respectively. For Ogbn-Products+ClusterGCN at $60\%$ sparsity, GST outperforms DSpar and LSim by $3.45\%$ and $7.95\%$, respectively.

\textbf{Obs.\blackcircle{3} Different GNN backbones and graphs show varying resistance to sparsification.} As shown in Fig.~\ref{fig:rq1_1}, GIN/GAT are less affected by graph sparsification compared to GCN. At $80\%$ graph sparsity, sparsifiers generally lead to an accuracy drop of  $7.08\%\sim20.7\%$ on GCN+Cora, whereas that on GAT+Cora is only $6.44\%\sim14.1\%$. Moreover, the resilience to sparsification varies across graphs; specifically, GAT+Citeseer experiences a $3.67\%\sim19.9\%$ decline at $80\%$ sparsity, while GAT+PubMed shows only a $3.16\%\sim8.46\%$ degradation.

\begin{figure}[!ht]
\centering
\includegraphics[width=1\columnwidth]{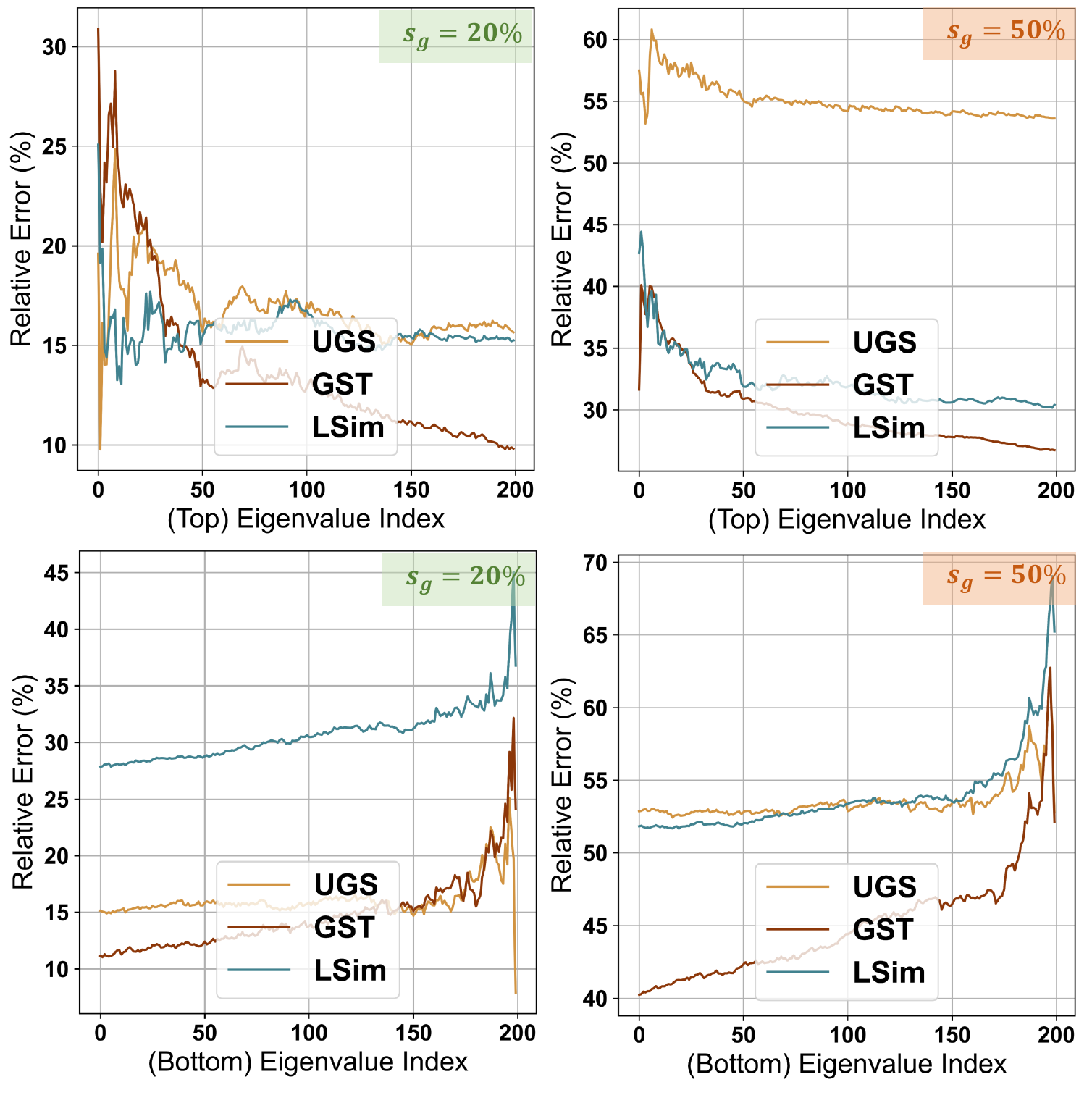}
\vspace{-2.2em}
\caption{The relative error of the top-200 and bottom-200 eigenvalues on PubMed+GCN, \emph{i.e.}, $\frac{\lambda_i - \lambda'_i}{\lambda_i}$, sparsified by different methods at sparsity level $20\%$ and $50\%$.}
\vspace{-1em}
\label{fig:rq2_1}
\end{figure}

\vspace{-0.3em}
\subsection{Spectral Preservation Helps Sparsification ($\mathcal{RQ}$2)}
\label{sec:topo}
\vspace{-0.3em}
To answer $\mathcal{RQ}$2, we compare the eigenvalue variation between sparse and original graphs, following \cite{liu2023dspar}. In \Cref{fig:rq2_1}, we showcase the relative error in the top-200 and bottom-200 eigenvalues of sparse graphs produced by sparsifiers compared to the original graph, on Citeseer/PubMed+GIN/GAT. We select only the top/bottom-200 eigenvalues because small (large) eigenvalues represent the global clustering (local smoothness) structures of the graphs, adequately reflecting the preservation of spectral information~\cite{liu2023dspar}. We observe that:

\textbf{Obs.\blackcircle{4} GST effectively preserves key eigenvalues.} As demonstrated in \Cref{fig:rq2_1,fig:rq2_2,fig:rq2_3,fig:rq2_4}, for the top-200 eigenvalues, GST achieves a performance similar to or better than LSim, and significantly surpasses UGS at $50\%$ sparsity. On Citeseer+GIN, its relative error fluctuates only within $0\%\sim8\%$. Regarding the bottom-200 eigenvalues, GST provides the best approximation. At $20\%$ sparsity, its average relative error is $20\%$ lower than that of UGS. More analyses can be found in Appendix ~\ref{app:spectral}.

\begin{figure}[!ht]
\centering
\includegraphics[width=0.8\columnwidth]{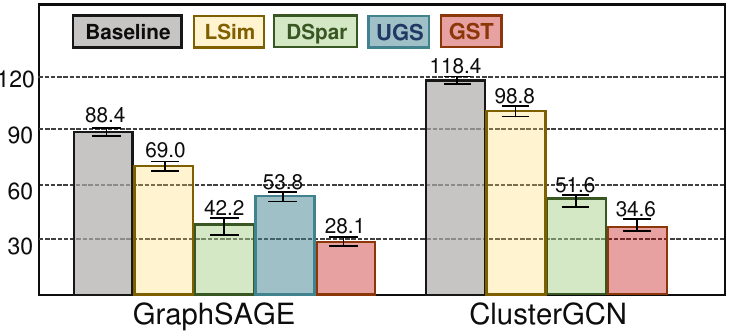}
\vspace{-1em}
\caption{The inference latency on Ogbn-Proteins with different sparsifiers when their performance loss is negligible ($\leq 1\%$).}
\vspace{-1em}
\label{fig:rq3_1}
\end{figure}

\begin{table*}[!t]
\centering
\caption{Efficiency comparison between GST and other sparsifiers. ``Sparsity (\%)'' indicates the extreme graph sparsity of different sparsifiers; ``MACs (m)'' represents the inference MACs ($=\frac{1}{2}$FLOPs); ``Latency (ms)'' refers to GPU reference latency. We \colorbox{gray!25}{shade} the best results and \underline{underline} the second-best results for each dataset. The GPU latency results are averaged over five random trials.}
% \vspace{-0.1em}
\setlength\tabcolsep{7pt}
\resizebox{1\textwidth}{!}{
\begin{tabular}{lccccccccccccccc}
\toprule\midrule % & \multirow{2}{*}{Methods}
\multicolumn{2}{c}{\multirow{2}{*}{Model}} & \multicolumn{3}{c}{\large
 Cora} & \multicolumn{3}{c}{\large Citeseer} & \multicolumn{3}{c}{\large PubMed}  & \multicolumn{3}{c}{\large Avg.}  \\
\cmidrule(lr){3-5} \cmidrule(lr){6-8}\cmidrule(lr){9-11}\cmidrule(lr){12-14}
 & & \makecell{Sparsity (\%)}& {\makecell{MACs (m)}} & \makecell{Latency (ms)}& \makecell{Sparsity (\%)}& {\makecell{MACs (m)}} & \makecell{Latency (ms)} &\makecell{Sparsity (\%)}& {\makecell{MACs (m)}} & \makecell{Latency (ms)}&Rank&\makecell{MAC Savings} & Speedup\\
\midrule
\parbox[t]{2mm}{\multirow{3}{*}{\rotatebox[origin=c]{90}{GCN}}} & Baseline & -- & $1996.55$ & $4.27$ & -- & $6318.00$ & $6.08$ & -- & $5077.84$ & $6.25$ & -- & $0\%$ & $1.00\times$  \\
 & SCAN & $11.93\%$ & $1758.37$ & $4.13$ & $6.52\%$ & $5526.99$ & $5.88$ & $23.68\%$ & $3875.41$ & $5.52$ & $\mathbf{7}$ & $16.04\%$ & $1.03\times$  \\
 & LSim & $11.58\%$ & $1765.35$ & $3.99$ & $21.83\%$ & $4938.78$ & $5.35$ & $11.84\%$ & $4476.62$ & $5.67$ & $\mathbf{5}$ & $15.06\%$ & $1.10\times$  \\
  & DSpar & $9.78\%$ & $1882.44$ & $4.20$ & $8.22\%$ & $5943.22$ & $5.97$ & $18.60\%$ & $4266.18$ & $5.49$& $\mathbf{6}$ & $8.65\%$ & $1.05\times$\\
 & UGS & $19.10\%$ & $1758.37$ & $4.13$& $27.20\%$ & $4599.50$ & $5.22$ & $24.91\%$ & {$3905.67$} & {$5.12$} & $\mathbf{3}$ & $23.73\%$ & $1.13\times$ \\
 & WD-GLT& $19.46\%$ & $1702.34$ & $4.24$ & $19.46\%$ & $4989.16$& $5.37$ & $11.24\%$ & $4455.10$ & $5.67$ & $\mathbf{4}$ & $18.20\%$ & $1.11\times$ \\
  & MGSpar & $31.00\%$ & \underline{$1568.21$} & \underline{$3.92 (1.08\times)$} & $30.00\%$ & \underline{$4318.75$} & \underline{$5.03 (1.20\times)$} & $27.00\%$ & \underline{$3758.53$} & \underline{$5.11(1.22\times)$} & $\mathbf{2}$ & $24.67\%$ & $1.13\times$  \\
 & \ourmethod & $40.00\%$ & \cellcolor{gray!25}$1397.59$ & \cellcolor{gray!25}$3.36 (1.27\times)$& $43.00\%$ & \cellcolor{gray!25}$3890.80$ & \cellcolor{gray!25}$4.48 (1.35\times)$& $35.00\%$ &\cellcolor{gray!25} $3092.81$ & \cellcolor{gray!25}$4.82 (1.29\times)$& $\mathbf{1}$ & 34.49\% & $1.30\times$ \\
\midrule

\parbox[t]{2mm}{\multirow{3}{*}{\rotatebox[origin=c]{90}{GIN}}} & Baseline & -- & $2006.26$ & $2.53$ & -- & $6328.22$ & $5.02$ & -- & $5108.12$ & $6.12$ & -- & $0\%$ & $1.00\times$\\

 & SCAN & $11.93\%$ & $1766.91$ & $2.24$ & $12.52\%$ & $5535.93$ & $3.84$ & $11.82\%$ & $4504.34$ & $5.81$ & $\mathbf{5}$ & $12.10\%$ & $1.16\times$\\
 & LSim & $11.58\%$ & $1773.93$ & $2.34$ & $10.43\%$ & $5668.19$ & $4.11$ & $11.84\%$ & $4503.32$ & $5.74$ & $\mathbf{6}$ & $11.28\%$ & $1.12\times$\\
  & DSpar & $21.11\%$ & \underline{$1533.29$} & \underline{$2.02 (1.25\times)$}& $14.77\%$ & $5372.09$ & $3.71$ & $16.18\%$ & $4157.10$ & $5.16$& $\mathbf{3}$ & $19.09\%$ & $1.26\times$\\
 & UGS & $19.4\%$ & $1617.05$ & $2.18$ & $27.5\%$ & \cellcolor{gray!25}$4587.96$ & \underline{$3.60(1.39\times)$}& $27.6\%$ & $3698.28$ & $5.02$ & $\mathbf{4}$ & $24.83\%$ & $1.25\times$  \\
 & WD-GLT & $19.7\%$ & $1590.33$ & $2.16$ & $16.4\%$ & $5198.11$ & $3.73$& $32.4\%$ & \underline{$3450.59$} & \underline{$4.89 (1.25\times)$} & $\mathbf{2}$ & $23.66\%$ & $1.27\times$  \\
 & MGSpar & $10.00\%$ & $1784.52$ & $2.39$ & $5.00\%$ & $6087.18$ & $4.90$& $7.00\%$ & $4885.16$ & $5.94$ & $\mathbf{7}$ & $6.49\%$ & $1.05\times$ \\
 & \ourmethod & $31.00\%$ & \cellcolor{gray!25}$1304.27$ & \cellcolor{gray!25}$1.85 (1.36\times)$& $26.43\%$ & \underline{$4633.97$} & \cellcolor{gray!25}$3.52 (1.42\times)$ & $53.00\%$ & \cellcolor{gray!25}$3109.47$ & \cellcolor{gray!25}$4.57 (1.33\times)$ & $\mathbf{1}$& $33.29\%$ & $1.37\times$\\
\midrule

\parbox[t]{2mm}{\multirow{3}{*}{\rotatebox[origin=c]{90}{GAT}}} & Baseline & -- & $8029.60$ & $13.2$ & -- & $25309.91$ & $13.5$ & -- & $20675.00$ & $15.3$ & -- & $0\%$ & $1.00\times$ \\
 & SCAN & $24.16\%$ & $6089.65$ & $9.9$ & $24.87\%$ & $19015.33$ & $11.8$ & $23.68\%$ & \underline{$15779.16$} & \underline{$12.6(1.21\times)$} & $\mathbf{3}$ & $22.23\%$ & $1.23\times$  \\
 
 & LSim & $23.22\%$ & $6165.13$ & $10.2$ & $21.83\%$ & $19784.75$ & $11.8$ & $11.84\%$ & $18227.08$ & $14.9$& $\mathbf{6}$ & $16.96\%$ & $1.15\times$  \\
 
  & DSpar & $17.34\%$ & $6788.29$ & $11.33$& $22.85\%$ & $19655.05$ & $11.7$ & $11.58\%$ & $18106.99$ & $14.8$ & $\mathbf{7}$ & $15.25\%$ & $1.12\times$ \\
  
 & UGS & $18.22\%$ & $6584.27$ & $11.15$ & $25.54\%$ & $18850.75$ & $12.36$ & $18.22\%$ & $16953.50$ & $13.92$ & $\mathbf{6}$ & $18.66\times$ & $1.13\times$ \\
 
 & WD-GLT & $19.71\%$ & \underline{$6504.33$} & \underline{$10.91(1.20\times)$} & $28.54\%$ & $18223.13$ & $11.57$ & $13.54\%$ & $17784.61$ & $14.50$ & $\mathbf{4}$ & $18.59\%$ & $1.16\times$  \\
 
  & MGSpar & $19.00\%$ & $6400.25$ & $11.08$& $32.00\%$ & \underline{$17114.76$}& \underline{$11.19(1.20\times)$}& $20.00\%$ & $16309.54$ & $12.8$& $\mathbf{2}$ & $27.59\%$ & $1.23\times$ \\
 & \ourmethod & $35.00\%$ & \cellcolor{gray!25}$5216.28$ & \cellcolor{gray!25}$7.90 (1.67\times)$& $41.00\%$ & \cellcolor{gray!25}$14920.45$& \cellcolor{gray!25}$10.30(1.31\times)$ & $39.00\%$ & \cellcolor{gray!25}$13438.75$ & \cellcolor{gray!25}$11.23 (1.36\times)$& $\mathbf{1}$ & $37.22\%$ & $1.45\times$ \\
\midrule
\bottomrule
\end{tabular}}\label{tab:efficiency}
\vspace{-0.9em}
\end{table*}

\vspace{-0.3em}
\subsection{GST Significantly Accelerates Computations ($\mathcal{RQ}$3)}
\vspace{-0.4em}
To answer $\mathcal{RQ}$3, we exhaustively compare the efficiency of different sparsifiers via three metrics, extreme graph sparsity, inference MACs, and GPU inference latency in \Cref{tab:efficiency} and \Cref{fig:rq3_1,fig:rq3_2}. Further explanations on these metrics can be found in Appendix~\ref{app:eff}.  We give the following observations:

\textbf{Obs.\blackcircle{5} GST significantly accelerates GNN inference.} More specifically, GST's inference speedup is more pronounced for large-scale graphs compared to smaller ones. As shown in \Cref{tab:efficiency}, GST provides $33.29\%\sim37.22\%$ MACs savings for GCN/GIN/GAT, with average inference speedups of $1.30\times$, $1.37\times$, and $1.45\times$, respectively. On Ogbn-Proteins, this can reach $3.14\sim3.42\times$ (in \Cref{fig:rq3_1}). In conclusion, GST significantly aids in accelerating GNN inference at a practically applicable level.

\vspace{-0.5em}
\subsection{High Robustness and Versatility of GST ($\mathcal{RQ}$4)}\label{sec:rq4}
\begin{figure}[!t]
\centering
\includegraphics[width=1\columnwidth]{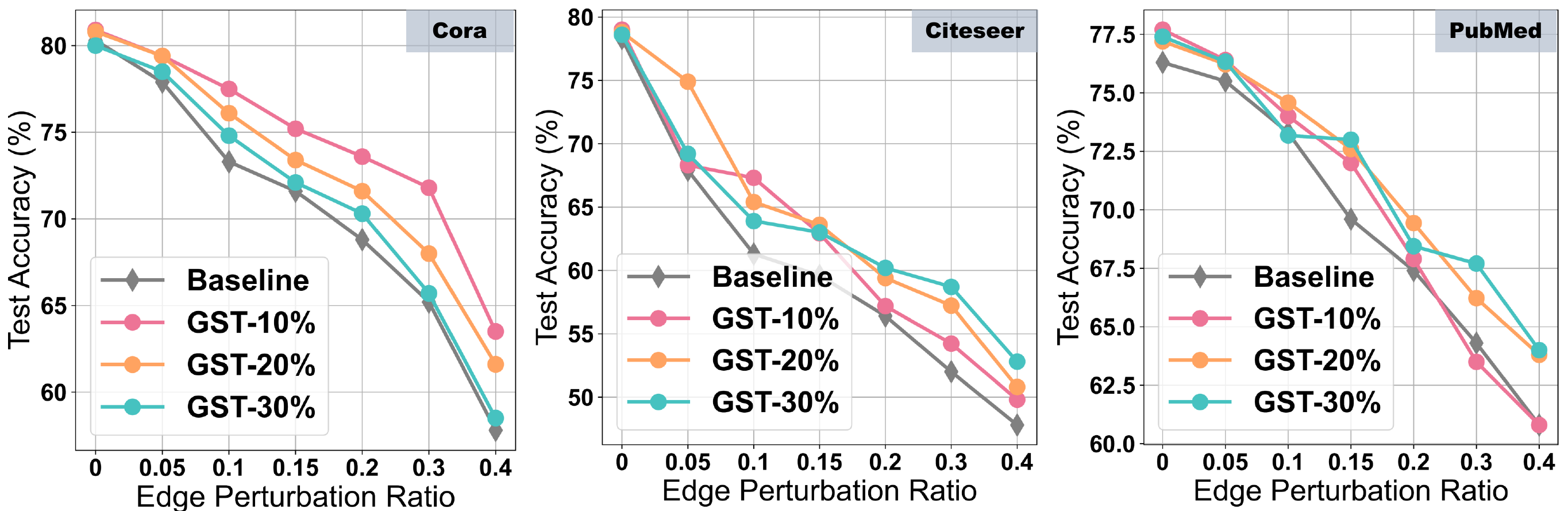}
\vspace{-2em}
\caption{The robust performance of GST on edge perturbations with a varying fraction of perturbed edges ($0\%\rightarrow40\%$).}
\vspace{-1.8em}
\label{fig:rq4_1}
\end{figure}

\vspace{-0.4em}
To validate the versatility of GST, this section examines (1) its robustness against edge perturbations, and (2) its applicability to graph lottery tickets identification. In the perturbation experiments, we induce edge perturbations by randomly relocating edge endpoints. For the graph lottery ticket experiment, we simply replace the graph mask found by UGS~\cite{chen2021unified} with that by GST in the graph lottery ticket and retrain the GNN from scratch. \Cref{fig:rq4_1,fig:rq4_defense} illustrate the performance of GST at various sparsity levels against edge perturbation, and \Cref{tab:rq4_ugs} presents the performance improvement achieved by combining GST with UGS. Observations include: \blackcircle{1} GST significantly enhances GNN's robustness against edge perturbations. For instance, on Cora+GCN, GST at $10\%$ sparsity achieved up to $7.23\%$ performance improvement (in \Cref{fig:rq4_defense}). \blackcircle{2} Across all sparsity levels, GST consistently aids UGS in identifying graph lottery tickets (in \Cref{tab:rq4_ugs}).
Detailed analysis is provided in Appendix \ref{app:versa}.

\vspace{-0.5em}
\subsection{Ablation \& Parameter Sensitivity Analysis ($\mathcal{RQ}$5)}
\label{sec:ablation}
\vspace{-0.4em}
\textbf{Ablation Study.} In order to better verify the effectiveness of each component of
GST, We have made corresponding modifications to GST and designed the following three variants: (1) \textbf{GST \textit{w/o} tuning}: moving the dynamic sparse training part of GST; (2) \textbf{GST \textit{w/o} sema}: merely utilizing $\phi^{(\text{sema})}$ when updating the sparse graph; (3) \textbf{GST \textit{w/o} topo}: merely utilizing $\phi^{(\text{topo})}$ when updating the sparse graph.

As indicated in \Cref{tab:rq5_ablation,fig:rq5_ablation}, it is evident that \blackcircle{1} the removal of the dynamic fine-tuning process in GST \textit{w/o} tuning leads to a significant performance drop; \blackcircle{2} using either $\phi^{(\text{topo})}$ or $\phi^{(\text{sema})}$ alone cannot match the performance of the original GST, with the omission of $\phi^{(\text{sema})}$ having a more pronounced impact. In summary, removing either component deteriorates the effectiveness of GST. For detailed data, as well as parameter sensitivity experiments, refer to Appendix \ref{app:rq5}.

\vspace{-1.2em}
\begin{table}[ht]\footnotesize
\centering
\caption{Ablation study on GST with its three variants. We report the extreme graph sparsity on Citeseer+GCN/GIN and Ogbn-ArXiv+GraphSAGE/ClusterGCN.}
\vspace{-0.1em}
\setlength{\tabcolsep}{1.80pt}
\resizebox{\columnwidth}{!}{
\begin{tabular}{l|cc|cc}
\toprule
Dataset&  \multicolumn{2}{c}{Citeseer}   &  \multicolumn{2}{c}{Ogbn-ArXiv} \\
% \cmidrule(r){1-1}\cmidrule(r){2-5}
\midrule
Backbone& GCN & GIN & GraphSAGE   & ClusterGCN \\
% \cmidrule(r){1-1}\cmidrule(r){2-3}\cmidrule(r){4-5}
\midrule
GST & $43.43_{\pm1.46}$ &$26.43_{\pm1.07}$ &  $35.18_{\pm0.72}$ &  $21.44_{\pm0.95}$ \\
\midrule
GST \textit{w/o} tuning& $30.41_{\pm0.75}$ &$10.39_{\pm1.78}$ &  $18.32_{\pm1.14}$ &  $12.57_{\pm0.56}$ \\
GST \textit{w/o} sema & $39.82_{\pm1.49}$ &$25.09_{\pm1.32}$ &  $29.78_{\pm1.42}$ &  $19.24_{\pm0.85}$ \\
GST \textit{w/o} topo & $42.08_{\pm0.56}$ &$25.93_{\pm1.26}$ &  $32.21_{\pm0.49}$ &  $19.67_{\pm0.79}$ \\
\bottomrule
\end{tabular}}
\label{tab:rq5_ablation}
\vspace{-1.2em}
\end{table}

\vspace{-0.5em}
\section{Conclusion \& Future Work}
\vspace{-0.5em}
This paper studies the notorious inefficiency of GNNs. Different from previous research literature, we open a novel topic, termed Graph Sparse Training (GST), for the first time. GST aims to fine-tune the sparse graph structure during the training process, utilizing anchors derived from full graph training as supervisory signals. GST also proposes the {Equilibria Principle} to balance both topological and semantic information preservation. Extensive experiments demonstrate that integrating the GST concept can enhance both performance and inference speedup. Additionally, GST can serve as a philosophy to benefit a wide array of training algorithms. In the future, we plan to extend GST to more complex scenarios, including heterophilic and heterogeneous graphs, and further explore the feasibility of in-time sparsification \& acceleration in real-world applications (recommender systems, fraud detection, \textit{etc.}).

\bibliography{example_paper}
\bibliographystyle{icml2024}
%%%%%%%%%%%%%%%%%%%%%%%%%%%%%%%%%%%%%%%%%%%%%%%%%%%%%%%%%%%%%%%%%%%%%%%%%%%%%%%
%%%%%%%%%%%%%%%%%%%%%%%%%%%%%%%%%%%%%%%%%%%%%%%%%%%%%%%%%%%%%%%%%%%%%%%%%%%%%%%
% APPENDIX
%%%%%%%%%%%%%%%%%%%%%%%%%%%%%%%%%%%%%%%%%%%%%%%%%%%%%%%%%%%%%%%%%%%%%%%%%%%%%%%
%%%%%%%%%%%%%%%%%%%%%%%%%%%%%%%%%%%%%%%%%%%%%%%%%%%%%%%%%%%%%%%%%%%%%%%%%%%%%%%
\newpage
\appendix
\onecolumn

\section{Notations}

\begin{table}[htbp]\footnotesize
  \centering
  \caption{The notations that are commonly used in Methodology (Sec.~\ref{sec:method}).}
   \setlength{\tabcolsep}{16pt} %biao kuan 
   \renewcommand\arraystretch{1.3} %
  \vspace{1em}
    \begin{tabular}{cc}
    \toprule
    Notation & Definition \\
    \midrule
    $\mathcal{G} = \left\{\mathcal{V}, \mathcal{E} \right\} = \left\{  \mathbf{A},\mathbf{X}\right\}$          & Input graph \\
    $\mathbf{A}$  & Input adjacency matrix \\
    $\mathbf{X}$  & Node features \\
    $\mathbf{\Theta}$ & Weight matrices of GNN \\
    $\lambda_i(\mathcal{G})$  & The $i$-th eigenvalue of $\mathcal{G}$ \\
     $s_g$ & Target graph sparsity \\
     $\Psi$ & Graph masker \\
     $\boldsymbol{m}_g^{\mathcal{A}}$ & Graph mask at the point of the highest validation score during full-graph training\\
     $\mathbf{Z}^{\mathcal{A}}=f(\{\boldsymbol{m}^{\mathcal{A}}_g\odot\mathbf{A},\mathbf{X}\},\mathbf{\Theta}^\mathcal{A})$ & GNN output at the point of the highest validation score during full-graph training \\
     $\mathcal{G}^{\mathcal{A}}=\{\boldsymbol{m}_g^{\mathcal{A}}\odot\mathbf{A},\mathbf{Z}^{\mathcal{A}} \}$ & Anchor graph that contains the topological \& semantic information of the original graph \\

    $\mathbf{M}_g^{\text{OS}}$ & (Binary) one-shot graph mask \\

    $\mathbf{M}_g^{(\mu)}$ & (Binary) graph mask at $\mu$-th interval \\

    $\mathcal{E}_{(\mu)}$ & Remained edges at $\mu$-th interval \\

    $\mathcal{E}_{(\mu)}^C$ & Pruned edges at $\mu$-th interval \\

    \bottomrule
    \end{tabular}%
  \label{tab:notation}%
\end{table}%

\section{More Dicussions on Related Work}\label{app:related_work}

% \subsection{Related Works on Graph Sparsification}\label{app:gspar}

% \subsection{Comparison with Dynamic Sparse Training}\label{app:dst_comp}

% 为了更突出的体现我们的贡献，在本节我们将会详细阐述our proposed GST 与传统dynamic sparse training方法的异

\begin{table*}[ht!]
\scriptsize
\setlength\tabcolsep{0.2pt}
\resizebox{\columnwidth}{!}{\begin{tabular}{l|ccc|ccc}
 \textbf{Sprase Training} & {\textbf{\small Target}} & \textbf{\small Semantic?} & \textbf{\small Topology?} & \textbf{\small PRC$^\S$} & \textbf{\small Criterion$\dag$} & \textbf{\small Backbone}  \\ \hline \hline
 {SNIP~\cite{lee2018snip}} & Sparse NN & \greencheck & \redcheck & \greencheck & magnitude, gradient & AlexNet, VGG, LSTM, \textit{etc.}  \\
{SET~\cite{mocanu2018scalable}} & Sparse NN & \greencheck & \redcheck & \greencheck & magnitude, gradient& MLP, CNN  \\
{SNFS~\cite{dettmers2019sparse}} & Sparse NN & \greencheck & \redcheck & \greencheck & magnitude, momentum & AlexNet, VGG, ResNet, \textit{etc.}  \\
{RigL~\cite{evci2020rigging}} & Sparse NN & \greencheck & \redcheck & \greencheck & magnitude, gradient & ResNet, MobileNet   \\
{ITOP~\cite{liu2021we}} & Sparse NN & \greencheck & \redcheck & \greencheck & magnitude, gradient & MLP, VGG, ResNet, \textit{etc.} \\
{DST~\cite{liu2020dynamic}} & Sparse NN & \greencheck& \redcheck & \redcheck & magnitude  & LeNet, LSTM, VGG, \textit{etc.}\\
{MEST~\cite{yuan2021mest}} & Sparse NN & \greencheck& \redcheck & \greencheck & magnitude, gradient & ResNet \\
{IMDB~\cite{hoang2023revisiting}} & Sparse NN & \redcheck & \greencheck & \greencheck & IMDB property & ResNet, VGG  \\
{DSnT~\cite{zhang2023dynamic}} & Sparse LLM & \greencheck & \redcheck & \greencheck & magnitude, output variation & LLaMA, Vicuna, OPT   \\
\textbf{GST (Ours)} & Sparse Graph & \greencheck & \greencheck & \greencheck & magnitude, eigenvalue, gradient & GCN, GIN, GAT, \textit{etc.}  \\

 \hline \hline
\multicolumn{7}{l}{
  \begin{minipage}{16cm}
    \footnotesize $\S$ Prune Rate Control. Whether the method has control over the sparsity rate. \\ 
    \footnotesize $\dag$ Criterion refers to the drop or regrow criterion during connection update.\\
\end{minipage}
}\\
\end{tabular}}
\caption{Comparison among different sparse training techniques, regarding their sparsification target, semantic and topological awareness, sparsity rate controllability, drop/grow criterion and backbones. Note that our proposed GST is the initial endeavor to apply the concept of dynamic sparse training to graph data, incorporating graph-related domain knowledge. }
\label{tab:sparsifiers}

\end{table*}

To highlight our contributions, we detail in Tab.~\ref{tab:sparsifiers} how our proposed GST differs from traditional (dynamic) sparse training methods, with the two most crucial distinctions as follows:
\begin{itemize}
    \item \textbf{Sparsification Target}: Previous dynamic sparse training methods primarily focused on pruning parameters, typically within traditional CNN frameworks (\textit{e.g.}, VGG, Resnet) or large language models (\textit{e.g.}, LLaMA, OPT). In contrast, GST is a novel exploration of dynamically sparsifying the input data, specifically graph data.
    \item \textbf{Drop \& Regrow Criterion}: Most prior sparse training techniques utilize semantic information from the model training procedure (\textit{e.g.}, magnitude, momentum, gradients) for parameter pruning. GST is not a trivial adaptation of DST to graph data; instead, it represents the first attempt that combines well-developed semantic criteria and domain knowledge about graph topology. Specifically, GST, for the first time, considers both the impact of sparsification on model output (reflecting semantics) and on graph Laplacian eigenvalues (reflecting topology). It optimizes these two objectives in tandem to obtain the optimal sparse subgraph.
\end{itemize}

\section{Eigenvalue Variation Approximation}\label{app:eigen}

Given the anchor graph $\mathcal{G}^{\mathcal{A}} = \{ \boldsymbol{m}^{\mathcal{A}}\odot\mathbf{A},\mathbf{Z}^\mathcal{A} \}$, we aim to assess how removing a specific edge affects $\mathcal{G}^{\mathcal{A}}$'s overall topological properties, as indicated by the eigenvalue variation, as follows:
\begin{equation}
\begin{aligned}
\phi^{\text{(topo)}}(e_{ij}) = \sum_{k=1}^{N} \frac{|\Delta \lambda_k|}{\lambda_k} = \sum_{k=1}^{N} \frac{|\lambda_k(\mathcal{G}^{\mathcal{A}}) - \lambda_k(\mathcal{G}^{\mathcal{A}} \setminus e_{ij})|}{\lambda_k(\mathcal{G}^{\mathcal{A}})},
\end{aligned}
\end{equation}
However, recalculating the $N$ eigenvalues for each $\mathcal{G}^\mathcal{A}\setminus e_{ij}$ theoretically requires $\mathcal{O}(E\cdot N^3)$ complexity, which is computationally impractical. Therefore, we consider approximating the eigenvalue variation. 

We denote the impact of removing $e_{ij}$ on the anchor graph as $\Delta\mathbf{A}$, where $\Delta\mathbf{A}_{ij}=-1$, and the rest are zeros. The anchor graph after removing $e_{ij}$ can be expressed as $\boldsymbol{m}^\mathcal{A}_g\odot(\mathbf{A} + \Delta\mathbf{A})$. Similarly, the influence of removing $e_{ij}$ on the right eigenvalue $\lambda_k$ and the corresponding right eigenvector $b^{(k)}$ of $\mathcal{G}^\mathcal{A}$ is denoted as $\Delta\lambda_k$ and $\Delta b$, respectively. According to the definitions of eigenvalues and eigenvectors, we have:
\begin{equation}
\left(\boldsymbol{m}^\mathcal{A}_g\odot(\mathbf{A} + \Delta\mathbf{A})\right) (b^{(k)} + \Delta b^{(k)}) = (\lambda_k + \Delta \lambda_k) (b^{(k)} + \Delta b^{(k)}).
\end{equation}
It is noteworthy that for large matrices, it is reasonable to assume that the removal of a link or node has a minor impact on the spectral properties of the graph. Therefore, both $\Delta b^{(k)}$ and $\Delta \lambda^{(k)}$ are small. Left-multiplying this equation by the transpose of the left eigenvector $a^T$ and neglecting second-order terms $a^{(k)T}\Delta \mathbf{A} \Delta b^{(k)}$ and $\Delta \lambda_k a^{(k)T} \Delta b^{(k)}$, we have:
\begin{equation}
\Delta \lambda_k = \frac{\boldsymbol{m}_{g,ij}^\mathcal{A}a^{(k)}_ib^{(k)}_j}{a^{(k)T}b^{(k)}},
\end{equation}
Based on this, the eigenvalue variation of $e_{ij}$ can be further expressed as:
\begin{equation}
\phi^{\text{(topo)}}(e_{ij}) = \sum_{k=1}^{N} \left|\frac{\boldsymbol{m}_{g,ij}^\mathcal{A}a^{(k)}_ib^{(k)}_j}{\lambda_ka^{(k)T}b^{(k)}}\right|,
\end{equation}
However, for large graphs, computing all eigenvalues/eigenvectors still incurs significant computational overhead. Considering that small (large) eigenvalue can effectively indicate the global clustering (local smoothness) structure of the graphs~\cite{zhang2019prone}, we only select the top-$K$ and bottom-$K$ ($K$=20) eigenvalues to compute their variation:
\begin{equation}
\phi^{\text{(topo)}}(e_{ij}) =  (\sum_{k=1}^{K} + \sum_{k=N-K}^{N}) \left|\frac{\boldsymbol{m}_{g,ij}^\mathcal{A}a^{(k)}_ib^{(k)}_j}{\lambda_ka^{(k)T}b^{(k)}}\right|,
\end{equation}

Additionally, we utilize non-trivial approximations of eigenvalues/eigenvectors with sublinear, \textit{i.e.}, $o(N^2)$, time complexity~\cite{bhattacharjee2021sublinear} to reduce the computational budget further.

\section{Compelxity Analysis}\label{app:complexity}

During the full-graph training stage, the inference time complexity of GST is:
\begin{equation}
\mathcal{O}\left(\underbrace{L\times E \times D}_{\text{aggregation}} + \overbrace{E\times D}^{\text{graph masker}} + \underbrace{L\times N \times D}_{\text{update}}\right), 
\end{equation}
where $L$ is the number of GNN layers and $D$ is the feature dimension. Subsequently, we approximate the eigenvalue/eigenvector for each edge, with a sublinear time complexity of $o(N^2)$. It is worth noting that this computation is performed only once. The inference time complexity of the sparse training procedure is:
\begin{equation}
\mathcal{O}\left(\underbrace{L\times \| \mathbf{M}_g \odot \mathbf{A}\|_0 \times D}_{\text{sparsified aggregation}} + \overbrace{\| \mathbf{M}_g \odot \mathbf{A}\|_0\times D}^{\text{graph masker}} + \underbrace{L\times N \times D}_{\text{update}}\right),
\end{equation}
The memory complexity of GST is:
\begin{equation}
\mathcal{O}\left(\underbrace{L\times N \times D}_{\text{node embeddings}} + \overbrace{L \times |\mathbf{\Theta}| \times D^2}^{\text{GNN parameter}} + \underbrace{|\Psi| \times D}_{\text{graph masker}}\right)
\end{equation}

\section{Algorithm Framework}\label{app:algo}

The algorithm framework is presented in Algo.~\ref{alg:algo}.
\begin{algorithm}[!t]
\caption{Algorithm workflow of \ourmethod}\label{alg:algo}
\Input{$\mathcal{G}=(\mathbf{A},\mathbf{X})$, GNN model $f(\mathcal{G}, \mathbf{\Theta}_0)$, GNN's initialization $\mathbf{\Theta}_0$, target sparsity $s_g\%$, update interval $\Delta T$, the number of epochs to obtain anchor $E$, the number of epochs for sparse graph fine-tuning $D$, learning rate $\eta$.}
                
\Output{Sparse graph $\mathcal{G}^{\text{sub}} = \{\mathbf{M}_g\odot\mathbf{A},\mathbf{X}\}$}

\For{\rm{iteration} $i \leftarrow 1$ \KwTo $E$}{

Compute the edge mask $\boldsymbol{m}_g^i$ via graph masker $\Psi$ \Comment*[r]{\textcolor{blue}{Eq.~\ref{eq:masker}}}

Forward $f_{sub}\left( \{\boldsymbol{m}_g^i \odot \mathbf{A}, \mathbf{\Theta}_i \}, \boldsymbol{m}_\theta \right) $ to compute $\mathcal{L}_{\text{anchor}}$\Comment*[r]{\textcolor{blue}{Eq.~\ref{eq:loss}}}

Backpropagate to update $\mathbf{\Theta}_{i+1} \leftarrow \mathbf{\Theta}_{i} - \eta \nabla_{\mathbf{\Theta}}\mathcal{L}_{\text{anchor}}$.

Update masks $\boldsymbol{m}_g$ with graph masker $\Psi$. % \Comment*[r]{\textcolor{blue}{Eq.~\ref{eq:masker}}}
}
\tcc{\textcolor{blue}{Obtain Anchor Graph}}
Record the {anchor graph} $\mathcal{G}^\mathcal{A} = \{m_g^\mathcal{A}\odot \mathbf{A},\mathbf{Z}^{\mathcal{A}}\}$ with the highest validation score.

\tcc{\textcolor{blue}{Obtain One-shot Mask}}
Set $s_g\%$ of the lowest magnitude values in $\boldsymbol{m}_g^\mathcal{A}$ to 0 and others to 1, then obtain one-shot mask $\mathbf{M}^{\text{OS}}_g$.

\tcc{\textcolor{blue}{Dynamically Update Edge Mask}}

Set $\mathbf{M}_g^{(1)} \leftarrow \mathbf{M}^{\text{OS}}_g$.

\For{\rm{iteration} $d \leftarrow 1$ \KwTo $D$}{

Compute interval index $\mu \leftarrow \lceil d / \Delta T \rceil$.

Forward $f\left(  \{\boldsymbol{m}_g \odot \mathbf{M}^{(\mu)}_g \odot \mathbf{A}, \mathbf{X}\},   \mathbf{\Theta} \right)$ to compute the $\mathcal{L}_{\text{anchor}}$.

Update $\mathbf{\Theta}$ and $\boldsymbol{m}_g$ accordingly.

\tcc{\textcolor{blue}{Update Graph Structure}}
\uIf{ $\mu = \lceil d / \Delta T \rceil$}{

    Set $\mathcal{E}_{(\mu)} \leftarrow$ edges in $\mathbf{M}_g^{(\mu)}\odot\mathbf{A}$,\; $\mathcal{E}_{(\mu)}^{C} \leftarrow$ edges in $\neg\mathbf{M}_g^{(\mu)}\odot\mathbf{A}$.

    \For{\rm{edge} $(i,j)$ \rm{in} $\mathcal{E}$}{
    Compute semantic criteria $\phi^{(\text{sema})}(e_{ij})  \leftarrow \left|\nabla_{e_{ij}}\mathcal{S}(\mathcal{G}^\mathcal{A},\mathcal{G}^{(\mu)})\right|$. \Comment*[r]{\textcolor{blue}{Eq.~\ref{eq:sema}}}

    Compute topological criteria $\phi^{\text{(topo)}}(e_{ij})  \leftarrow \left(\sum_{k=1}^{K} + \sum_{k=N-K}^{N}\right)\frac{\boldsymbol{m}^{\mathcal{A}}_{g,ij}a^{(k)}_ib^{(k)}_j}{\lambda_k(\mathcal{G}^{\mathcal{A}})a^{(k)T}b^{(k)}}$ \Comment*[r]{\textcolor{blue}{Eq.~\ref{eq:topo}}}

    Combine semantic and topological criteria $\phi(e_{ij})  \leftarrow \beta^s\cdot\phi^{\text{(sema)}}(e_{ij}) + \beta^t\cdot\phi^{\text{(topo)}}(e_{ij})$\Comment*[r]{\textcolor{blue}{Eq.~\ref{eq:combine}}}
    }

    Compute the number of edges to be swapped $r \leftarrow \mathbf{\Upsilon(\mu)}$\Comment*[r]{\textcolor{blue}{Eq.~\ref{eq:scheduler}}}

\tcc{\textcolor{blue}{Select Drop/Regrow Edges}}
    Set $\mathbf{M}^{(\text{prune})} \leftarrow \operatorname{ArgTopK}\left(-\phi(\mathcal{E}_{(\mu)}), r\right),
\mathbf{M}^{(\text{regrow})} \leftarrow \operatorname{ArgTopK}\left(\phi(\mathcal{E}_{(\mu)}^C), \; r\right)$\Comment*[r]{\textcolor{blue}{Eq.~\ref{eq:choose}}}

    Update edge masks $\mathbf{M}_g^{(\mu+1)} \leftarrow \left( \mathbf{M}_g^{(\mu)} \setminus \mathbf{M}_g^{(\text{prune})} \right) \cup \mathbf{M}_g^{(\text{regrow})}$\Comment*[r]{\textcolor{blue}{Eq.~\ref{eq:update}}}

  }
}
\end{algorithm}

\section{Dataset Description}

We conclude the dataset statistics in Tab.~\ref{tab:dataset}

\begin{table}[htb]
\centering
\vspace{-4mm}
\caption{Graph datasets statistics.}
% \vspace{-0.5em}
\label{tab:dataset}
% \resizebox{\columnwidth}{!}{
\begin{tabular}{c  |c |c |c |c |c | c  }
\toprule
Dataset   &  Nodes & Edges & Ave. Degree & Features & Classes & Metric  \\ 
\midrule
Cora  & 2,708 &5,429 & 3.88 & 1,433  & 7 & Accuracy\\ 
Citeseer & 3,327 & 4,732 &1.10 & 3,703 & 6 & Accuracy \\ 
PubMed& 19,717 & 44,338 & 8.00 & 500  & 3 &  Accuracy \\ \midrule

Ogbn-ArXiv  & 169,343 & 1,166,243 & 13.77 & 128 & 40 & Accuracy \\
Ogbn-Proteins  & 132,534 & 39,561,252 & 597.00 & 8 & 2 & ROC-AUC \\
{Ogbn-Products} &{2,449,029} & {61,859,140} & {50.52} & {100} & {47} & {Accuracy} \\
\bottomrule
\end{tabular}%}
\vspace{-2mm}
\end{table}

\section{Experimental Configurations}\label{sec:app_setup}

\subsection{Train-val-test Splitting of Datasets.} To rigorously verify the effectiveness of our proposed GST, we unify the dataset splitting strategy across all GNN backbones and baselines. As for node classification tasks of small- and medium-size datasets, we utilize 420 (Citeseer) and 460 (PubMed) labeled data for training, 500 nodes for validation and 500 nodes for testing. For Squirrel and Chameleon datasets, we follow the original settings in~\cite{cui2023mgnn,rozemberczki2021multi}, and set the train/valid/test ratio as 60\%/20\%/20\%.  The data splits for Ogbn-ArXiv, Ogbn-Proteins, and Ogbn-Products were provided by the benchmark~\cite{hu2020open}. Specifically, for Ogbn-ArXiv, we train on papers published until 2017, validate on papers from 2018 and test on those published since 2019. For Ogbn-Proteins, protein nodes were segregated into training, validation, and test sets based on their species of origin. For Ogbn-Products, we use the sales ranking (popularity) to split nodes into training/validation/test sets. Concretely, we sort the products according to their sales ranking and use the top 8\% for training, the next top 2\% for validation, and the rest for testing. % For Ogbl-Collab, we employed collaborations until 2017 as training edges, those in 2018 as validation edges, and those in 2019 as test edges.

\subsection{Baseline Configurations} We detail how we report the results of baseline methods:

\begin{itemize}
    \item Topology-based sparsification
    \begin{itemize}
        \item \textbf{SCAN}~\cite{spielman2008graph}: SCAN uses structural similarity (called SCAN similarity) measures to detect clusters, hubs, and outliers. We utilize the implementation in~\cite{chen2023demystifying}
        \item \textbf{Local Similarity}~\cite{hamann2016structure}: Local Similarity ranks edges using the Jaccard score and computes $\log(\operatorname{rank}(e_{ij}))/\log(\operatorname{deg}(e_{ij}))$ as the similarity score, and selects edges with the highest similarity scores. We utilize the implementation in~\cite{chen2023demystifying}.
        \item \textbf{DSpar}~\cite{liu2023dspar}: DSpar is an extension of effective resistance sparsifier, which aims to reduce the high computational budget of calculating effective resistance through an unbiased approximation. We adopt their official implementation~\cite{liu2023dspar}.
    \end{itemize}
    \item Semantic-based sparsification
    \begin{itemize}
        \item \textbf{UGS}~\cite{chen2021unified}: We utilize the official implementation from the authors. Notably, UGS was originally designed for joint pruning of model parameters and edges. Specifically, it sets separate pruning parameters for parameters and edges, namely the weight pruning ratio $p_\theta$ and the graph pruning ratio $p_g$. In each iteration, a corresponding proportion of parameters/edges is pruned. For a fairer comparison, we set $p_\theta=0\%$, while maintaining $p_g=5\%$ (consistent with the original paper).
        \item \textbf{WD-GLT}~\cite{hui2022rethinking}: WD-GLT inherits the iterative magnitude pruning paradigm from UGS, so we also set $p_\theta=0\%, p_g=5\%$ across all datasets and backbones. The perturbation ratio $\alpha$ is tuned among $\{0,1\}$. Since no official implementation is provided, we carefully reproduced the results according to the original paper.
        \item \textbf{Meta-gradient sparsifier}~\cite{wan2021edge}: The Meta-gradient sparsifier prunes edges based on their meta-gradient importance scores, assessed over multiple training epochs. Since no official implementation is provided, we carefully replicated the results following the guidelines in the original paper.

    \end{itemize}
\end{itemize}

In addition to our selected baselines, we also enumerate other classic baselines relevant to graph sparsification and explain why they were not chosen for our study:

\begin{itemize}
    \item \textbf{Effective Resistance Sparsifier}~\cite{spielman2008graph}: This method is one of the most renowned spectral sparsifiers. However, due to its high time complexity ($\mathcal{0}(E\log(|\mathcal{V}|)^3)$), we opted for its approximate version, DSpar~\cite{liu2023dspar}.
    \item \textbf{DropEdge}~\cite{rong2019dropedge}: Though DropEdge also involves dropping edges during training, the dropping process is random across different training epochs. Thus, it is not capable of outputting a compact yet performant subgraph, and we therefore do not consider it for comparison.
    \item \textbf{NeuralSparse}~\cite{zheng2020robust}: NeuralSparse is a well-recognized method for graph sparsification. Nevertheless, it cannot regulate the ultimate graph sparsity ratio. Consequently, we do not take it into consideration.
    \item \textbf{FastGCN}~\cite{chen2018fastgcn}: FastGCN and other graph samplers~\cite{chen2017stochastic,ying2018graph} samples neighbors for each in-batch node with a certain probability. However, the sampling process is only for training and does not essentially output a sparse graph. 
\end{itemize}

\subsection{Update Scheduler}

$\mathbf{\Upsilon}(\mu)$ is the update scheduler which determines the number of edges to be swapped at each update. We simply adopt the Inverse Power~\cite{zhu2017prune}:
\begin{equation}\label{eq:scheduler}
\mathbf{\Upsilon}(\mu) = \tau (1 - \frac{\mu}{\lceil D / \Delta T \rceil})^\kappa,
\end{equation}
where $\tau$ denotes the initial ratio and $\kappa$ is the decay factor controlling how fast the ratio decreases with intervals.

\subsection{Parameter Configurations}\label{app:parameter}
The main parameters of GST include:
% \begin{enumerate}
$E$ (number of epochs to acquire the anchor graph),
$D$ (number of epochs to dynamically fine-tune the sparse graph), $\tau$ (the initial ratio of edges to swap), $\kappa$ (the decay factor of $\mathbf{\Upsilon}(\mu)$), $\Delta T$ (the update interval) and the learning rate.

We uniformly set $\tau$ to $0.3$ and $\kappa$ to $1$. Ablation experiments regarding them are available in Appendix \ref{app:rq5}. The other hyperparameters are listed in \Cref{tab:parameter}.

\vspace{-1em}
\begin{table}[h]
  \centering
  \setlength{\tabcolsep}{0.80pt}
  \caption{Hyper-parameter configurations.}
  % \vspace{-3mm}
  \resizebox{1\linewidth}{!}{
  \scriptsize
    \begin{tabular}{c|ccc ccc ccc cc cc cc}
    \toprule
    \multicolumn{16}{c}{\centering \scriptsize Computing Infrastructures: NVIDIA Tesla V100 (32GB)  Software Framework: Pytorch} \\
    \midrule
     \multirow{2}{*}{Param} & \multicolumn{3}{c}{Cora}  & \multicolumn{3}{c}{Citeseer}  & \multicolumn{3}{c}{PubMed} &  \multicolumn{2}{c}{ArXiv}& \multicolumn{2}{c}{Proteins}& \multicolumn{2}{c}{Products} \\
     % \midrule
    
     & GCN & GIN & GAT & GCN & GIN & GAT & GCN & GIN & GAT & SAGE & ClusterGCN & SAGE & ClusterGCN  & SAGE & ClusterGCN   \\
    \cmidrule(r){1-1} \cmidrule(r){2-4}\cmidrule(r){5-7} \cmidrule(r){8-10}\cmidrule(r){11-12} \cmidrule(r){13-14}\cmidrule(r){15-16}
    $E$ & 100 & 100 & 100 & 100 & 200 & 200 & 200 & 200 & 200 & 200 & 100 & 200 & 200 &200&200\\
    $D$ & 400 & 400 & 400 & 500 & 500 & 500 & 600 & 600 & 600&300 & 300 & 300&300&300&300 \\
    lr & 0.001 & 0.001 & 0.001 & 0.001 & 0.001 & 0.001 & 0.001 & 0.001 & 0.001&0.01& 0.01&0.01 &0.01&0.01&0.01\\
    $\Delta T$ & 20 & 20 & 20 & 20 & 20 & 20 & 20 & 20 & 20 & 3& 3& 3 &3 &3&3\\

    \bottomrule
    \end{tabular}%
}
  \label{tab:parameter}%
  % \vspace{-0.6cm}
\end{table}%

\subsection{Performance Metrics} Accuracy represents the ratio of correctly predicted outcomes to the total predictions made. The ROC-AUC (Receiver Operating Characteristic-Area Under the Curve) value quantifies the probability that a randomly selected positive example will have a higher rank than a randomly selected negative example. Hit@50 denotes the proportion of correctly predicted edges among the top 50 candidate edges.

\subsection{Efficiency Metrics}\label{app:eff}
To assess the efficiency of sparse graphs generated by various sparsifiers, we utilize two metrics: MACs (Multiply-Accumulate Operations) and GPU Inference Latency (ms). MACs theoretically represent the model's inference speed, based on FLOPs (Floating Point Operations Per Second). Although {\fontfamily{lmtt}\selectfont \textbf{SpMM}} is theoretically faster than {\fontfamily{lmtt}\selectfont \textbf{MatMul}} based on MACs/FLOPs, this advantage is not always realized in practice due to {\fontfamily{lmtt}\selectfont \textbf{SpMM}}'s random memory access pattern. To gauge the real-world applicability of our method, we additionally measure latency on GPUs in milliseconds (ms).

\section{Additional Experimental Results}

% small graph
\subsection{Experiments for $\mathcal{RQ}1$}

\begin{figure*}[!ht]
\centering
\includegraphics[width=1\textwidth]{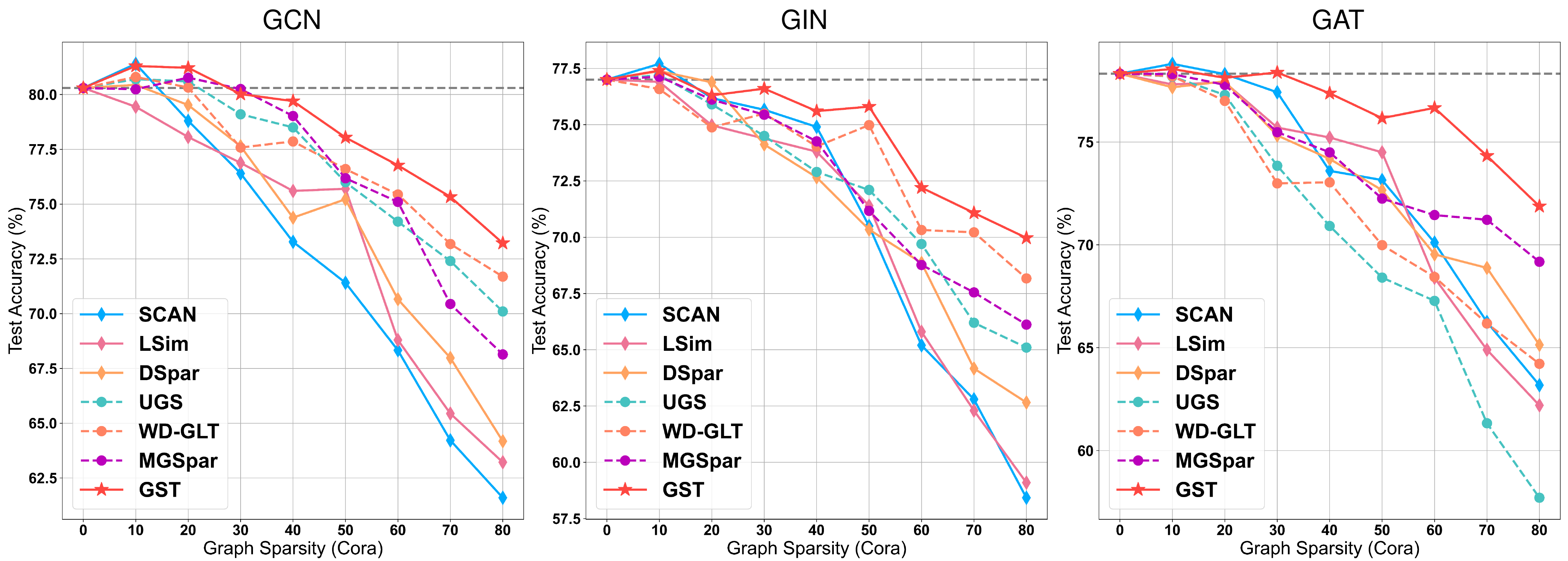}
\caption{Performance comparison of graph sparsification methods on Cora under different sparsity levels. The grey dashed line represents the original baseline performance.}
\label{fig:rq1_2}
\end{figure*}

This section details GST's performance on Cora, Ogbn-Proteins, and Ogbn-Products datasets. \Cref{fig:rq1_2} shows the performance of GST with GCN/GIN/GAT. Notably, GST excels on Cora+GAT, experiencing only a negligible performance loss ($\approx 1.8\%$) at $60\%$ graph sparsity.

\Cref{tab:rq1_proteins,tab:rq1_products} present GST's performance on Ogbn-Proteins and Products. LSim, DSpar, and UGS were chosen as baselines due to the limitations of other baselines in inductive settings or scalability to large graphs. Generally, GST maintains superior performance across all sparsity levels. Specifically, it preserves GraphSAGE/ClusterGCN performance with negligible loss ($\leq 1\%$) at $10\%\sim30\%$ sparsity. At $60\%$ sparsity, GST outperforms LSim/DSpar by up to $6.95\%$.

\vspace{-0.5em}
\begin{table}[!ht]\footnotesize
\centering
\caption{Performance comparison of sparsifiers at different sparsity levels ($10\%\rightarrow60\%$) on GraphSAGE/ClusterGCN with Ogbn-Proteins. We report the
mean accuracy ± stdev of 3 runs. We \colorbox{gray!25}{shade} the best results and \underline{underline} the second-best results.}
\vspace{0.4em}
\setlength{\tabcolsep}{1.80pt}
\resizebox{0.7\columnwidth}{!}{
\begin{tabular}{l|cccccc}
\toprule
 & \multicolumn{6}{c}{\textbf{GraphSAGE}}  \\ 
% \cmidrule(r){2-5}\cmidrule(r){6-9}
\cmidrule(r){1-1}\cmidrule(r){2-7}
 Sparsity& $10\%$ & $20\%$ & $30\%$  & $40\%$ & $50\%$ & $60\%$ \\
% \midrule
\cmidrule(r){1-1}\cmidrule(r){2-7}%\cmidrule(r){5-7}

LSim&  $76.92_{\pm0.58}$&  \underline{$75.03_{\pm1.17}$} &  \underline{$73.20_{\pm0.79}$} &  $73.69_{\pm0.52}$  &  $72.14_{\pm0.48}$  &  $70.09_{\pm0.59}$\\
DSpar &  $76.04_{\pm0.25}$&  \underline{$75.86_{\pm0.27}$} &  \underline{$74.46_{\pm0.39}$} &  $73.45_{\pm0.26}$  &  $70.22_{\pm0.54}$  &  $69.23_{\pm0.40}$\\

UGS &  $77.47_{\pm0.27}$&  $76.38_{\pm0.63}$ &  $74.58_{\pm0.49}$&  \underline{$72.12_{\pm0.38}$}  &  \underline{$72.38_{\pm0.33}$}  &  \underline{$71.45_{\pm0.75}$}\\

\ourmethod & \cellcolor{gray!25}{$77.59_{\pm0.52}$} &  \cellcolor{gray!25}$77.56_{\pm0.79}$  &  \cellcolor{gray!25}$76.45_{\pm0.66}$ &  \cellcolor{gray!25}$76.02_{\pm0.68}$ &  \cellcolor{gray!25}$75.45_{\pm0.77}$ &  \cellcolor{gray!25}$73.55_{\pm0.50}$\\
\midrule
 & \multicolumn{6}{c}{\textbf{ClusterGCN (GCN aggr)}
 } \\
 \cmidrule(r){1-1}\cmidrule(r){2-7}%\cmidrule(r){5-7}

 Sparsity & $10\%$ & $20\%$ & $30\%$   & $40\%$ & $50\%$ & $60\%$ \\
% \midrule
\cmidrule(r){1-1}\cmidrule(r){2-7}%\cmidrule(r){5-7}
LSim &  $75.44_{\pm0.57}$ &  $74.18_{\pm0.60}$  &  $74.27_{\pm0.89}$  &  $72.15_{\pm0.64}$ &  $69.42_{\pm0.91}$ &  $66.46_{\pm0.85}$\\
DSpar &  \underline{$76.38_{\pm0.79}$} &  \cellcolor{gray!25}$76.22_{\pm1.13}$  &  \underline{$75.28_{\pm0.85}$ } &  \underline{$74.66_{\pm1.05}$} &  \underline{$72.89_{\pm0.74}$ }&  \underline{$71.39_{\pm0.92}$}
\\
\ourmethod &  \cellcolor{gray!25}$76.29_{\pm0.56}$ &  \underline{$76.11_{\pm0.57}$}  &  \cellcolor{gray!25}$76.16_{\pm0.44}$ &  \cellcolor{gray!25}$75.83_{\pm0.69}$ &  \cellcolor{gray!25}$73.03_{\pm0.65}$  &  \cellcolor{gray!25}$73.80_{\pm0.77}$ \\
\bottomrule
\end{tabular}}
\label{tab:rq1_proteins}
\end{table}

\begin{table}[!ht]\footnotesize
\centering
\caption{Performance comparison of sparsifiers at different sparsity levels $\{10\%,20\%,30\%,40\%,50\%,60\%\}$ on GraphSAGE/ClusterGCN with Ogbn-Products. Due to the immense scale of Ogbn-Products, we report results from a single run only.  We \colorbox{gray!25}{shade} the best results and \underline{underline} the second-best results. }
\vspace{0.4em}
\setlength{\tabcolsep}{10pt}
\resizebox{0.7\columnwidth}{!}{
\begin{tabular}{l|cccccc}
\toprule
 & \multicolumn{6}{c}{\textbf{GraphSAGE}}  \\ 
% \cmidrule(r){2-5}\cmidrule(r){6-9}
\cmidrule(r){1-1}\cmidrule(r){2-7}%\cmidrule(r){5-7}
% \cmidrule(r){1-1}\cmidrule(r){2-5}
 % Ticket& \multicolumn{4}{c}{\textbf{Weight Sparsity}}  \\ 
 Sparsity& $10\%$ & $20\%$ & $30\%$  & $40\%$ & $50\%$ & $60\%$ \\
% \midrule
\cmidrule(r){1-1}\cmidrule(r){2-7}%\cmidrule(r){5-7}
LSim & $77.96$  & $76.60$ & $74.98$ & $72.23$ & $72.67$ & $72.49$\\
DSpar & \underline{$78.25$}  & \underline{$77.41$} & \underline{$75.19$} & \underline{$74.20$} & \underline{$74.57$} & \underline{$74.08$} \\
\ourmethod &  \cellcolor{gray!25}$78.79$ &  \cellcolor{gray!25}$78.52$  &  \cellcolor{gray!25}$77.15$ & \cellcolor{gray!25}$77.03$ & \cellcolor{gray!25}$75.86$ & \cellcolor{gray!25}$75.77$ \\
\midrule
 & \multicolumn{6}{c}{\textbf{ClusterGCN (GCN aggr)}} \\
 \cmidrule(r){1-1}\cmidrule(r){2-7}%\cmidrule(r){5-7}

 Sparsity & $10\%$ & $20\%$ & $30\%$   & $40\%$ & $50\%$ & $60\%$ \\
\cmidrule(r){1-1}\cmidrule(r){2-7}%\cmidrule(r){5-7}
LSim & $77.04$ & $74.34$ & $72.50$ & $70.24$ & $69.92$ & $65.10$ \\
DSpar & \underline{$78.25$} & \underline{$76.11$} & \underline{$74.39$} & \underline{$72.06$} & \underline{$69.74$} & \underline{$69.56$} \\
\ourmethod& \cellcolor{gray!25}$78.38$ & \cellcolor{gray!25}$78.45$ & \cellcolor{gray!25}$77.83$ & \cellcolor{gray!25}$77.19$ & \cellcolor{gray!25}$75.71$ & \cellcolor{gray!25}$73.05$  \\
\bottomrule
\end{tabular}}
\label{tab:rq1_products}
\end{table}

% spectral preservation
\subsection{Experiments for $\mathcal{RQ}2$}\label{app:spectral}

\begin{figure*}[!ht]
\centering
\includegraphics[width=0.6\textwidth]{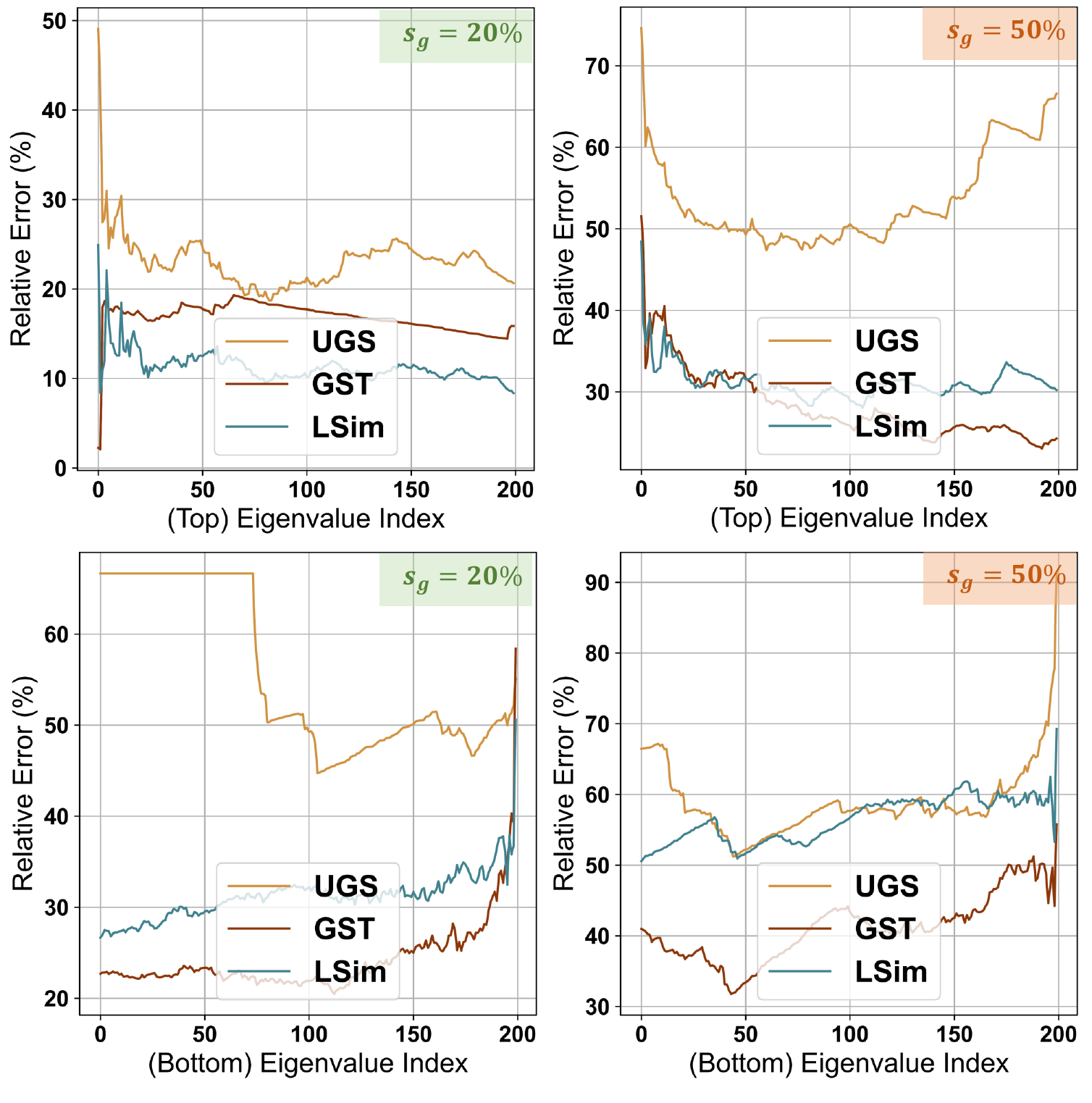}
\vspace{-0.5em}
\caption{The relative error of the top-200 and bottom-200 eigenvalues on Citeseer+GCN, \emph{i.e.}, $\frac{\lambda_i - \lambda'_i}{\lambda_i}$, sparsified by different methods at sparsity level $20\%$ and $50\%$.}
\label{fig:rq2_2}
\end{figure*}

\begin{figure*}[!ht]
\centering
\includegraphics[width=0.6\textwidth]{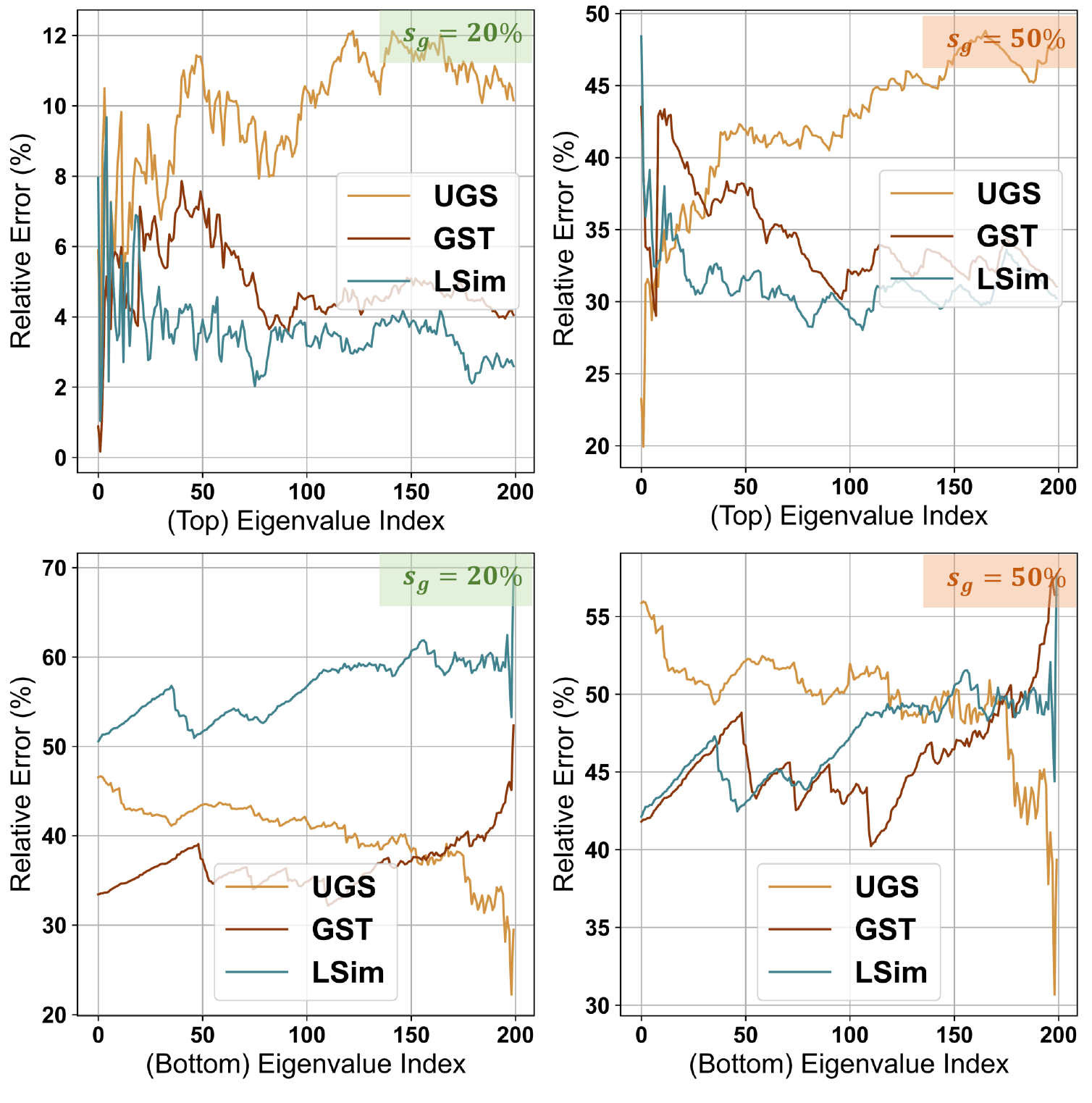}
\vspace{-0.5em}
\caption{The relative error of the top-200 and bottom-200 eigenvalues on Citeseer+GIN, \emph{i.e.}, $\frac{\lambda_i - \lambda'_i}{\lambda_i}$, sparsified by different methods at sparsity level $20\%$ and $50\%$.}
\label{fig:rq2_3}
\vspace{-1em}
\end{figure*}

\begin{figure*}[!ht]
\centering
\includegraphics[width=0.6\textwidth]{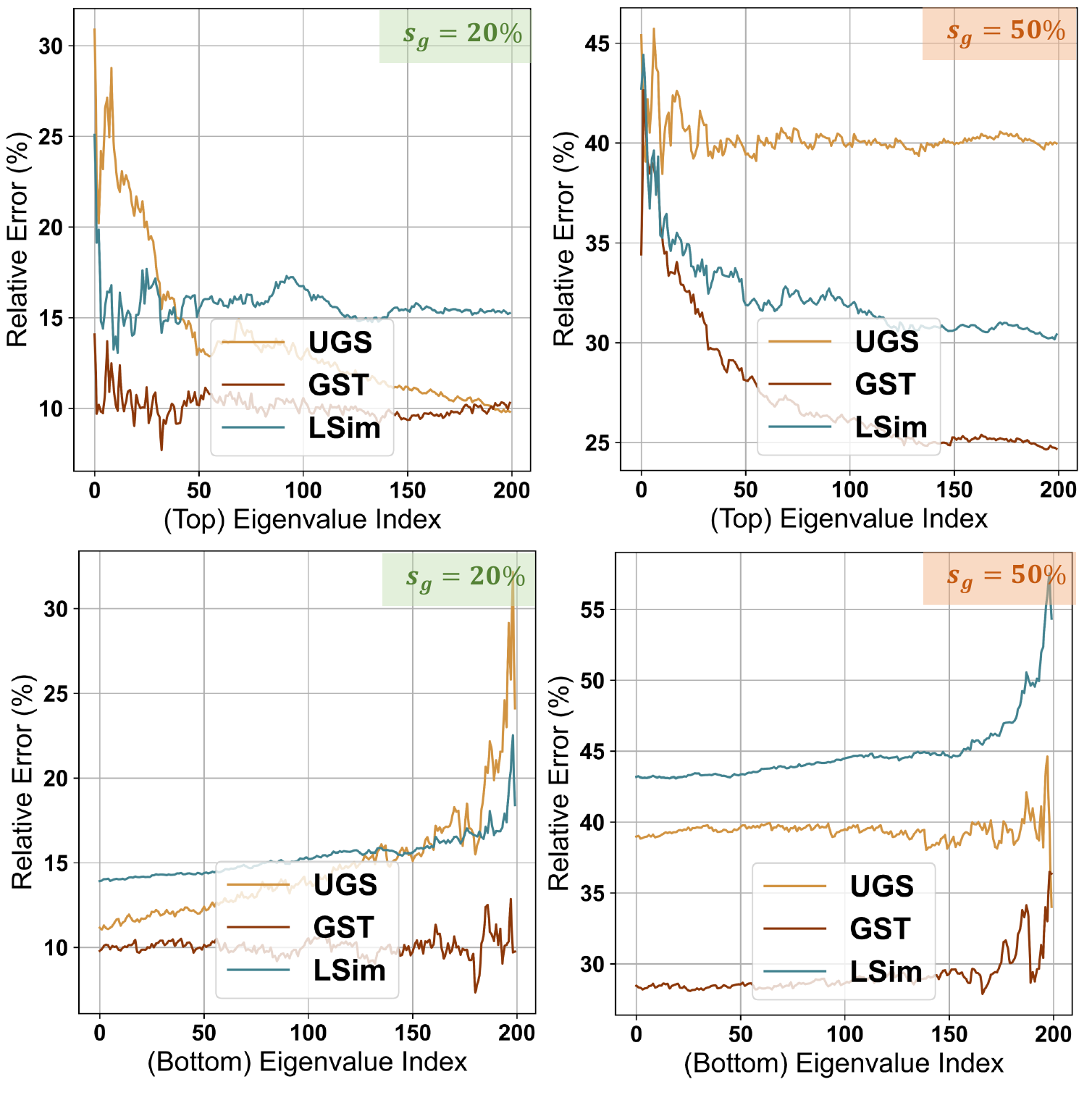}
\vspace{-0.5em}
\caption{The relative error of the top-200 and bottom-200 eigenvalues on PubMed+GIN, \emph{i.e.}, $\frac{\lambda_i - \lambda'_i}{\lambda_i}$, sparsified by different methods at sparsity level $20\%$ and $50\%$.}
\label{fig:rq2_4}
\end{figure*}

In \Cref{fig:rq2_1,fig:rq2_2,fig:rq2_3,fig:rq2_4}, we showcase the spectral preservation performance of GST, UGS, and LSim on Citeseer/PubMed with GCN/GIN. Specifically, we illustrate the distribution of relative error for the top-200 and bottom-200 eigenvalues at $20\%$ and $50\%$ sparsity levels produced by different methods. Our observations include: (1) GST significantly outperforms UGS/LSim in preserving the bottom-200 eigenvalues, which indicate local smoothness. For instance, on PubMed+GIN (in \Cref{fig:rq2_4}), at $50\%$ sparsity, GST's average relative error is about $10\%$ lower than UGS and $15\%$ lower than LSim, demonstrating GST's unique advantage in maintaining the local smoothness of sparse graphs. (2) For the top-200 eigenvalues, GST performs best on larger graphs, notably PubMed, with PubMed+GCN/GIN (in \Cref{fig:rq2_1,fig:rq2_4}) showing GST surpassing UGS/LSim for the top-200 eigenvalue preservation.

% inference speedup
\subsection{Experiments for $\mathcal{RQ}3$}

\begin{figure}[!t]
\centering
\includegraphics[width=1\columnwidth]{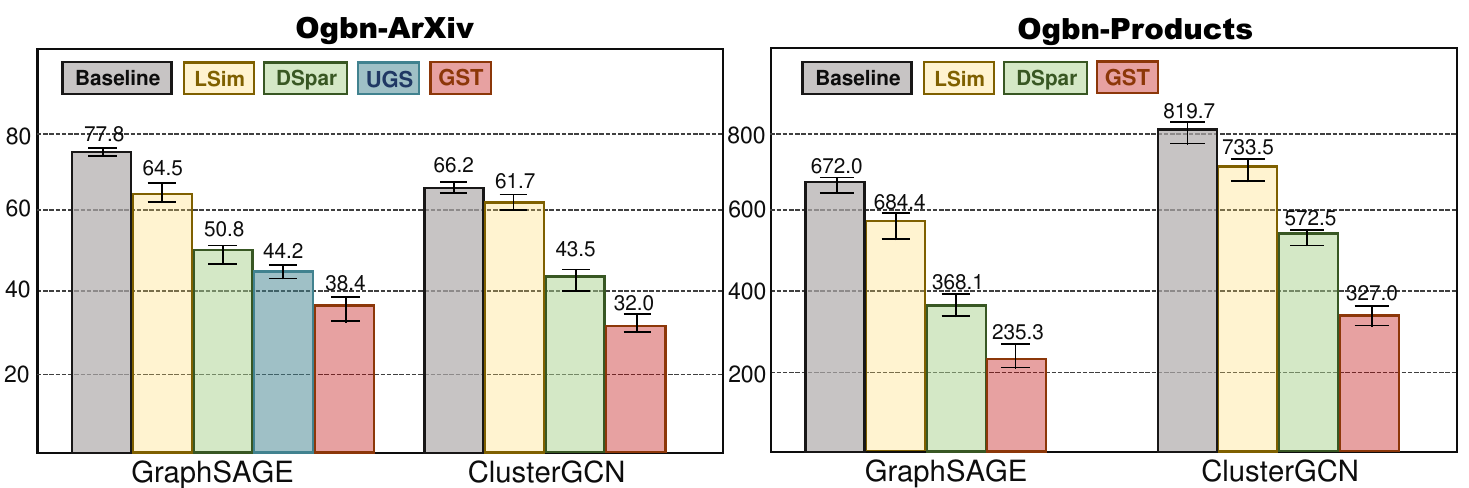}
\vspace{-2em}
\caption{The inference latency on Ogbn-ArXiv/Products with different sparsifiers when their performance loss is negligible ($\leq 1\%$).}
\vspace{-1em}
\label{fig:rq3_2}
\end{figure}

In \Cref{fig:rq3_1,fig:rq3_2}, we demonstrate GST's inference acceleration for GraphSAGE/ClusterGCN on OGB datasets. Due to UGS's inapplicability to the inductive ClusterGCN, values for UGS+ClusterGCN are omitted. It is observed that GST's inference acceleration on large-scale graphs is more pronounced than on small-scale graphs (as shown in \Cref{tab:efficiency}). With negligible performance loss ($\leq 1\%$), GST achieves an inference speedup of $2.50\sim2.85\times$ on Ogbn-Products and $3.14\sim3.42\times$ on Ogbn-Proteins. 

\subsection{Experiments for $\mathcal{RQ}4$}\label{app:versa}

As discussed in Sec.~\ref{sec:rq4}, we validate the versatility and plug-and-play nature of GST through two tasks: graph adversarial defense and graph lottery ticket identification. 

\begin{figure*}[!ht]
\centering
\includegraphics[width=1\textwidth]{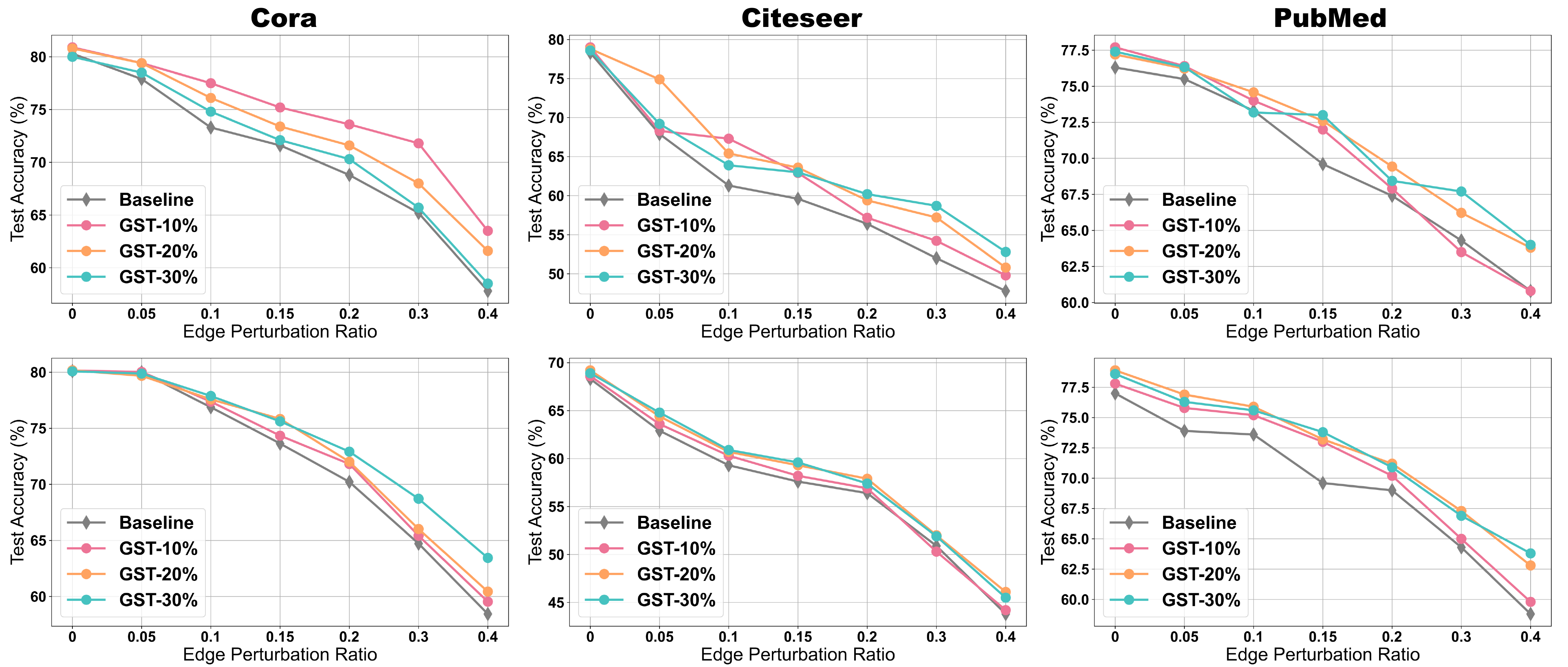}
\caption{The robust performance of GST on edge perturbations by perturbing a varying fraction of edges $\{0\%,5\%,10\%,15\%,20\%,30\%,40\%\}$, tested on GCN ($1^{\text{st}}$ row) and GAT ($2^{\text{nd}}$ row).}
\label{fig:rq4_defense}
\end{figure*}

\Cref{fig:rq4_defense} demonstrates how GST at sparsity levels of $\{10\%, 20\%, 30\%\}$ assists GCN/GAT in combating edge perturbations. Observations reveal that GST with proper sparsity significantly enhances GNN's resilience to edge perturbations. Across various datasets and backbones, the right level of graph sparsity notably improves GNN performance post-disturbance. For instance, on Cora+GCN, GST-$10\%$ recovers $7.23\%$ of GCN's performance under a $30\%$ perturbation ratio; on PubMed+GAT, GST-$30\%$ helps GAT regain $5.09\%$ performance at a $40\%$ perturbation ratio. This suggests that graphs of different sizes require varying degrees of sparsity for effective edge perturbation resistance.

\Cref{tab:rq4_ugs} presents the performance when replacing UGS's iteratively pruned sparse graphs with GST-discovered graphs of the same sparsity. The results show that in most iterations, GST-discovered graphs significantly outperform UGS-located graph lottery tickets, demonstrating GST's broad applicability.

\vspace{-0.5em}
\begin{table}[!ht]\footnotesize
\centering
\caption{Performance comparison of original UGS and UGS+GST on GCN backbone.}
\vspace{0.4em}
\setlength{\tabcolsep}{2.2pt}
% \resizebox{0.8\columnwidth}{!}{
\begin{tabular}{l|ccccccc}
\toprule

% \cmidrule(r){2-5}\cmidrule(r){6-9}
% \cmidrule(r){1-1}\cmidrule(r){2-7}
 Weight Sparsity& $20.0\%$ & $48.80\%$ & $67.23\%$  & $79.03\%$ & $86.58\%$ & $96.50\%$ & $98.95\%$ \\
 Graph Sparsity& $5.0\%$ & $14.3\%$ & $22.6\%$  & $30.17\%$ & $36.98\%$ & $43.12\%$ & $48.67\%$ \\
\midrule
Dataset & \multicolumn{6}{c}{\textbf{Cora}}  \\ 
% \cmidrule(r){1-1}\cmidrule(r){2-7}%\cmidrule(r){5-7}
\midrule
UGS &  $79.38$&  $78.08$ &  $77.36$&  {$78.05$}  &  {$76.22$}  &  $75.35$ &  $74.83$\\

\textbf{UGS+GST} & {$80.25$} &  $79.21$ & $78.83$ & $77.66$  & $77.24$ & $76.38$ & $75.62$\\
$\Delta$ & $0.87\uparrow$ & $1.13\uparrow$ & $1.47\uparrow$ & $0.39\downarrow$ & $1.02\uparrow$ & $1.03\uparrow$ & $0.79\uparrow$\\
\midrule
Dataset & \multicolumn{6}{c}{\textbf{Citeseer}
 } \\
% \cmidrule(r){1-1}\cmidrule(r){2-7}
\midrule
UGS &  $70.30$&  $69.77$ &  $68.14$&  {$69.02$}  &  {$68.53$}  &  $67.59$ &  $66.84$\\
% 70.5,70.5,71,70.6,69.8,69.2,67.8,66.2,64.7
\textbf{UGS+GST} & {$70.08$} &  $69.54$ & $69.48$ & $69.27$  & $68.79$ & $68.16$ & $67.34$\\

$\Delta$ & $0.22\downarrow$ & $0.23\downarrow$ & $1.34\uparrow$ & $0.22\uparrow$ & $0.82\uparrow$ & $0.57\uparrow$ & $0.5\uparrow$\\

\midrule
Dataset & \multicolumn{6}{c}{\textbf{PubMed}
 } \\
% \cmidrule(r){1-1}\cmidrule(r){2-7}
\midrule
UGS &  $78.51$&  $77.21$ &  $75.60$&  {$75.17$}  &  {$74.85$}  &  $75.21$ &  $74.96$\\
% 78.6,79.2,77.3,77,77.6,77.1,76.4,75.5,74.1]
\textbf{UGS+GST} & {$78.60$} &  $78.05$ & $77.18$ & $77.03$  & $75.92$ & $75.69$ & $75.30$\\
$\Delta$ & $0.09\uparrow$ & $0.84\uparrow$ & $1.58\uparrow$ & $0.86\downarrow$ & $1.07\uparrow$ & $0.48\uparrow$ & $0.34\uparrow$\\

\bottomrule
\end{tabular}
\label{tab:rq4_ugs}
\end{table}

% additional tasks
\subsection{Experiments for $\mathcal{RQ}5$}
\label{app:rq5}

\begin{figure*}[!ht]
\centering
\includegraphics[width=1\textwidth]{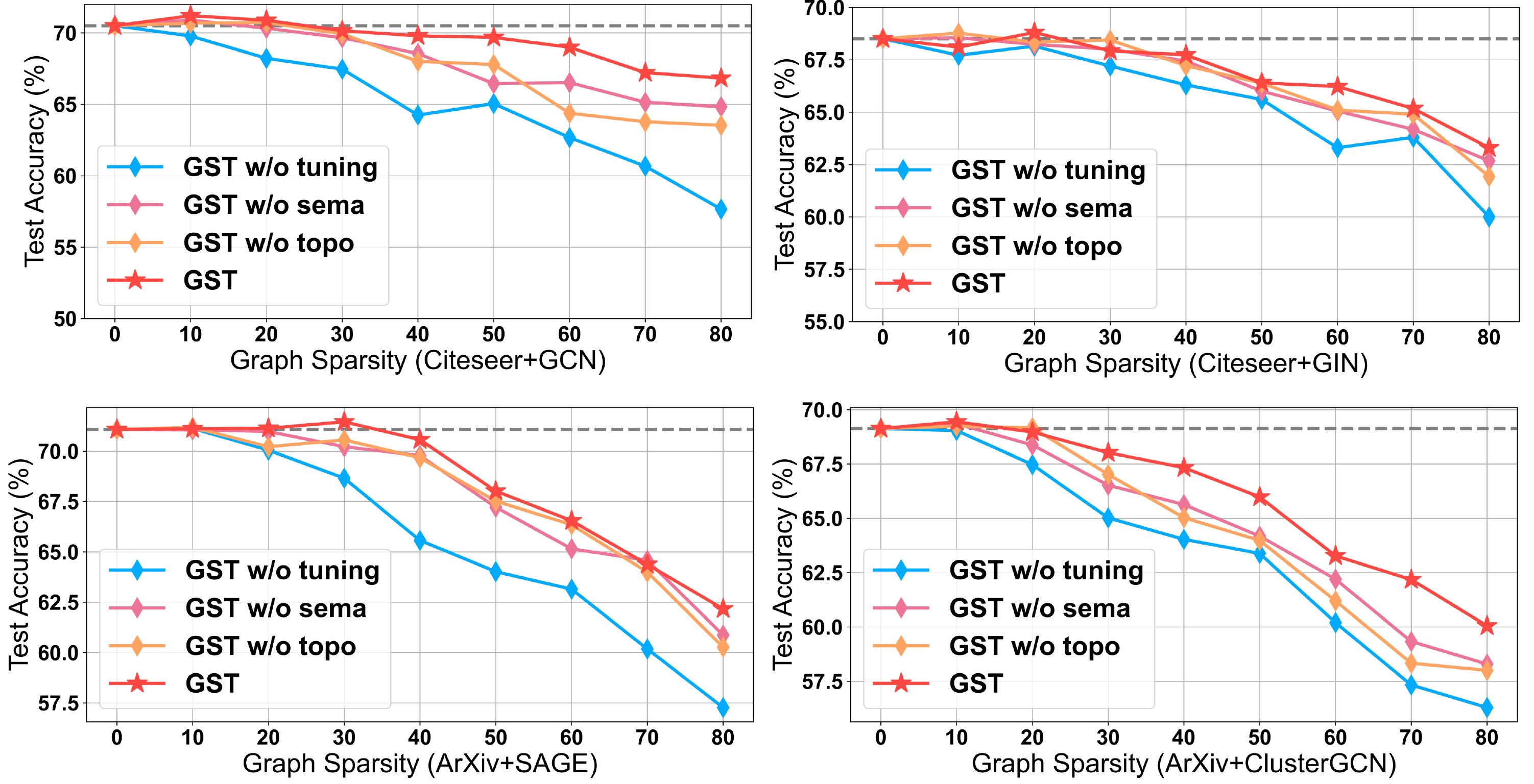}
\caption{Ablation study of GST tested on Citeseer+GCN/GIN and Ogbn-ArXiv+GraphSAGE/ClusterGCN(GCN aggr).}
\label{fig:rq5_ablation}
\end{figure*}

In this section, we present detailed data for ablation study and parameter sensitivity analysis. \Cref{fig:rq5_ablation} displays the performance of GST and its three variants on Citeseer+GCN/GIN and Ogbn-ArXiv+GraphSAGE/ClusterGCN. \Cref{tab:rq5_schedule} shows the performance of GST under various settings of $\alpha$ and $\kappa$, while \Cref{tab:rq5_delta} demonstrates how the performance of GST varies with $\Delta T$. 

Regarding $\Delta T$, we can observe that smaller graphs (Citeseer) benefit from a lower update frequency, whereas larger graphs (Ogbn-ArXiv) perform better with more frequent updates. Specifically, $\Delta T=20$ shows consistently good performance on Citeseer+GCN/GIN, and $\Delta T=3$ excels on Ogbn-Arxiv. This observation is intuitive: larger graphs, with their more complex structures and a wider variety of potential sparse graph structures, necessitate more frequent structural updates. Therefore, we standardize the setting of $\Delta T=20$ for small graphs and $\Delta T=3$ for large graphs.  

Regarding parameters $\tau$ and $\kappa$, it's clear that GST's performance is relatively insensitive to these choices, with fluctuations on GCN, GIN, and GAT not exceeding $1.69\%$, $1.88\%$, and $1.25\%$ respectively. Overall, however, a larger $\alpha$ and a smaller $\kappa$, which both correspond to more frequent structural updates, tend to yield better performance.

% \vspace{-1.2em}
\begin{table}[ht]\footnotesize
\centering
\caption{Ablation study on GST with its different $\Delta T$. We report the extreme graph sparsity on Citeseer+GCN/GIN and Ogbn-ArXiv+GraphSAGE/ClusterGCN.}
\vspace{-0.1em}
\setlength{\tabcolsep}{1.80pt}
% \resizebox{\columnwidth}{!}{
\begin{tabular}{l|cccc|cccc}
\toprule
Dataset&  \multicolumn{4}{c}{Citeseer}   &  \multicolumn{4}{c}{Ogbn-ArXiv} \\
% \cmidrule(r){1-1}\cmidrule(r){2-5}
\midrule
Backbone& \multicolumn{2}{c}{GCN} & \multicolumn{2}{c}{GIN} & \multicolumn{2}{c}{GraphSAGE}   & \multicolumn{2}{c}{ClusterGCN} \\
% \cmidrule(r){1-1}\cmidrule(r){2-3}\cmidrule(r){4-5}
\midrule
Sparsity & $20\%$ &$50\%$ &  $20\%$ &  $50\%$   & $20\%$ &$50\%$ &  $20\%$ &  $50\%$ \\
\midrule
$\Delta T=3$ & $69.16$ &$68.01$ &  $69.15$ &  $66.02$ & $71.80$ &$68.02$ &  $69.21$ &  $67.33$ \\
$\Delta T=5$ & $70.62$ &$68.73$ &  $69.52$ &  $66.05$ & $71.18$ &$67.95$ &  $69.10$ &  $67.84$ \\
$\Delta T=20$ & $70.89$ &$68.46$ &  $69.71$ &  $66.16$ & $68.04$ &$67.77$ &  $69.40$ &  $66.25$ \\
$\Delta T=50$ & $69.75$ &$67.59$ &  $68.48$ &  $65.43$ & $68.06$ &$66.08$ &  $69.23$ &  $66.90$ \\

\bottomrule
\end{tabular}
\label{tab:rq5_delta}
% \vspace{-1.0em}
\end{table}

% \vspace{-0.6em}
\begin{table}[ht]\footnotesize
\centering
\caption{Parameter sensitivity analysis on initial swapping ratio $\alpha$ and decay factor $\kappa$. We report the performance of GST on Citeseer dataset at $20\%$ graph sparsity.}
\vspace{0.4em}
\setlength{\tabcolsep}{1.80pt}
% \resizebox{1\columnwidth}{!}{
\begin{tabular}{l|ccccccccc}
\hline
\hline
\multirow{2}{*}{$20\%$} & \multicolumn{3}{c}{GCN} & \multicolumn{3}{c}{GIN} & \multicolumn{3}{c}{GAT}  \\ 
\cmidrule(r){2-4} \cmidrule(r){5-7} \cmidrule(r){8-10}
& $\alpha=0.1$& $\alpha=0.2$& $\alpha=0.3$ & $\alpha=0.1$& $\alpha=0.2$& $\alpha=0.3$ & $\alpha=0.1$& $\alpha=0.2$& $\alpha=0.3$  \\ 
% \cmidrule(r){2-4}
\midrule
%  Ticket& \multicolumn{3}{c}{\textbf{Weight Sparsity}}  \\ 
% \midrule
$\kappa=1$  & $69.32$ &  $70.86$  & $70.89$  & $67.64$ &  $68.98$  & $69.35$ & $68.03$ &  $68.42$  & $68.55$ \\
$\kappa=2$ &  $69.14$ &  $70.84$  &  $69.85$  & $67.25$ &  $70.86$  & $70.89$ & $67.66$ &  $68.12$  & $68.33$ \\
$\kappa=3$ &  $68.26$ &  $68.30$  &  $69.20$  &  $67.48$ &  $69.11$  &  $69.20$ & $67.50$ &  $67.17$  & $67.95$\\

\bottomrule
\bottomrule
\end{tabular}
\label{tab:rq5_schedule}
\vspace{-0.4em}
\end{table}

\end{document}